\documentclass[10pt,journal,compsoc]{IEEEtran}
\usepackage{graphicx}
\usepackage{amsmath,amssymb} 

\usepackage[pagebackref=true,breaklinks=true,letterpaper=true,colorlinks,bookmarks=false]{hyperref}

\usepackage{times}
\usepackage{helvet}
\usepackage{courier}
\usepackage{graphicx}
\usepackage{bm}
\usepackage{amsmath,amssymb} 
\usepackage{color}
\usepackage{xcolor}
\usepackage[normalem]{ulem}
\usepackage{bbm}
\usepackage{epstopdf}
\usepackage{caption}
\usepackage{subcaption}
\usepackage{enumitem}
\usepackage{calc}
\usepackage{multirow}
\usepackage{xspace}
\usepackage{booktabs}
\usepackage{mathrsfs}
\usepackage{array}
\usepackage[]{algorithm2e}
\usepackage[font=small,skip=4pt]{caption}
\usepackage{comment}
\newcommand{\figref}[1]{Fig\onedot~\ref{#1}}
\newcommand{\equref}[1]{Eq\onedot~\eqref{#1}}
\newcommand{\secref}[1]{Sec\onedot~\ref{#1}}

\newcommand{\ve}[1]{{\mathbf #1}} 
\newcommand{\hua}[1]{{\mathcal #1}}
\newcommand{\scr}[1]{{\mathcal #1}}

\newcommand{\thickhline}{%
    \noalign {\ifnum 0=`}\fi \hrule height 1pt
    \futurelet \reserved@a \@xhline
}

\newcommand{\yang}[1]{\textcolor{black}{#1}}
\newcommand{\chenxu}[1]{\textcolor{black}{#1}}

\DeclareRobustCommand\onedot{\futurelet\@let@token\@onedot}
\def\onedot{\ifx\@let@token.\else.\null\fi\xspace}
\def\eg{\emph{e.g.}}

\def\ie{\emph{i.e.}}

\def\etc{\emph{etc}\onedot}

\def\etal{\emph{et al.}}
\usepackage{hyperref}

\ifCLASSOPTIONcompsoc
  \usepackage[nocompress]{cite}
\else
  \usepackage{cite}
\fi

\ifCLASSINFOpdf
\else
\fi
\begin{document}
%
\title{Every Pixel Counts ++: Joint Learning of Geometry and Motion with 3D  Holistic Understanding}
%
%

\author{Chenxu~Luo*,
        ~Zhenheng~Yang*,
        ~Peng~Wang*,
        ~Yang~Wang,
        ~Wei~Xu,~\IEEEmembership{Member,~IEEE}
        Ram Nevatia,~\IEEEmembership{Fellow,~IEEE}
        and~Alan~Yuille,~\IEEEmembership{Fellow,~IEEE}
\IEEEcompsocitemizethanks{
\IEEEcompsocthanksitem P. Wang, Y. Wang and W. Xu are with Baidu Research Institution. C. Luo and A. Yuille are with Johns Hopkins University. Z. Yang and R. Nevatia are with University of Southern California. This project was done when C. Luo was an intern at Baidu. C. Luo and A. Yuille are partially supported by ONR with grant N00014-19-S-B001.}
\thanks{* Equal contribution. Corresponding author: P. Wang}}

\IEEEtitleabstractindextext{%
\begin{abstract}

Learning to estimate 3D geometry in a single frame and optical flow from consecutive frames by watching unlabeled videos via deep convolutional network has made significant progress recently. Current state-of-the-art (SoTA) methods treat the two tasks independently. One typical assumption of the existing depth estimation methods is that the scenes contain no independent moving objects. while object moving could be easily modeled using optical flow.
In this paper, we propose to address the two tasks as a whole, i.e. to jointly understand per-pixel 3D geometry and motion. This eliminates the need of static scene assumption and enforces the inherent geometrical consistency during the learning process, yielding significantly improved results for both tasks. We call our method as ``Every Pixel Counts++'' or ``EPC++''.
Specifically, during training, given two consecutive frames from a video, we adopt three parallel networks to predict the camera motion (MotionNet), dense depth map (DepthNet), and per-pixel optical flow between two frames (OptFlowNet) respectively.
The three types of information, are fed into a holistic 3D motion parser (HMP), and per-pixel 3D motion of both rigid background and moving objects are disentangled and recovered.
Various loss terms are formulated to jointly supervise the three networks. An effective adaptive training strategy is proposed to achieve better performance and more efficient convergence.
Comprehensive experiments were conducted on  datasets with different scenes, including driving scenario (KITTI 2012 and KITTI 2015 datasets), mixed outdoor/indoor scenes (Make3D) and synthetic animation (MPI Sintel dataset). Performance on the five tasks of depth estimation, optical flow estimation, odometry, moving object segmentation and scene flow estimation shows that our approach outperforms other SoTA methods, demonstrating the effectiveness of each module of our proposed method. Code will be available at: \href{https://github.com/chenxuluo/EPC}{https://github.com/chenxuluo/EPC}.
\end{abstract}

\begin{IEEEkeywords}
Depth Estimation, Optical Flow Prediction, Unsupervised Learning.
\end{IEEEkeywords}}

\maketitle

\IEEEdisplaynontitleabstractindextext

%
\IEEEpeerreviewmaketitle


%
%
%
%



\section{Introduction}
\label{sec:introduction}

\IEEEPARstart{E}{stimating} 3D geometry (\eg~per-pixel depth) from a single image, understanding motion (\eg~relative camera pose and object motion) and optical flow between consecutive frames from a video are fundamental problems in computer vision. They enable a wide range of real-world applications such as augmented reality~\cite{NewcombeLD11}, video analysis~\cite{tsai2016video,yang2017spatio} and robotics navigation~\cite{menze2015cvpr,desouza2002vision}.
In this paper, we propose an effective learning framework that jointly estimates per-pixel depth, camera motion and optical flow, using only unlabeled videos as training data. 



\begin{figure*}
\includegraphics[width=\textwidth]{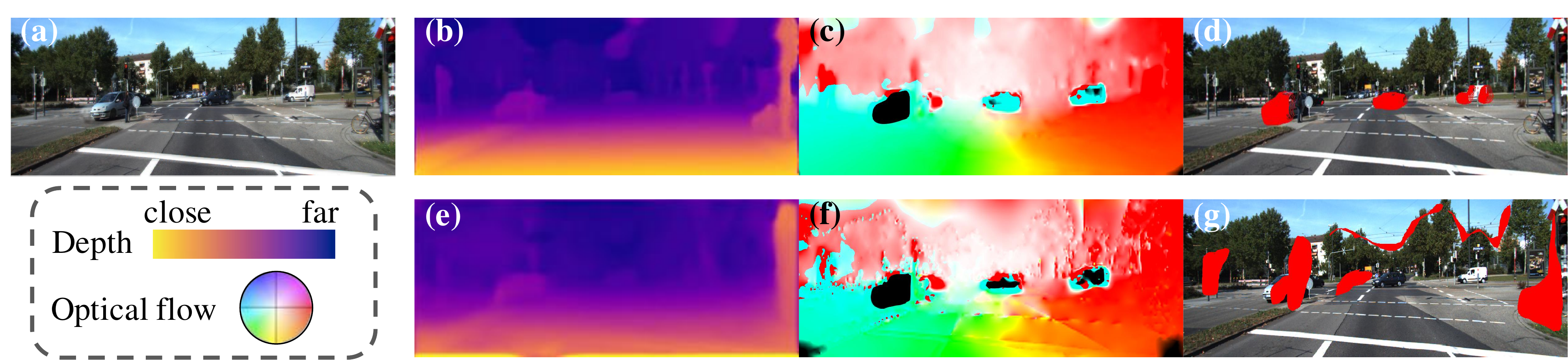}
\caption{(a) Two consecutive images (transparent second frame overlapped onto the first frame), 
(b) our estimated depth for the first frame,
(c) our estimated optical flow, (d) our estimated moving object mask, (e) depth from LEGO~\cite{yang2018cvpr} for the first frame, (f) optical flow estimation from Wang \etal~\cite{wang2017occlusion}, (g) segmentation mask from EPC~\cite{yang2018every}. By jointly learning all three geometrical cues, our results show significant improvement over other SoTA methods on different tasks.}
\label{fig:example}
\end{figure*}

Our work is motivated by recent unsupervised single image depth estimation approaches~\cite{godard2016unsupervised,zhou2017unsupervised,yang2018aaai,yang2018cvpr} which train a depth estimation deep network taking unlabeled video frames as input and using supervision with view-synthesis. Their estimated depths are even better than results from those of some supervised methods~\cite{eigen2014depth} in outdoor scenarios.
Specifically, the core idea follows the rule of rigid structure from motion (SfM)~\cite{wu2011visualsfm}, where the image of one view (source image) is warped to another view (target image) based on the predicted depth map of target image and their relative 3D camera motion. The photometric error between the warped image and the target image is used to supervise the learning of networks. 

However, real world videos may contain moving objects, which are inconsistent with the rigid scene assumption used in the earlier frameworks. 
Zhou \etal~\cite{zhou2017unsupervised} try to avoid such errors by inducing an explanability mask, where both pixels from moving objects and occluded regions are ignored during training.
Vijayanarasimhan \etal~\cite{Vijayanarasimhan17} separately tackle moving objects with a multi-rigid body model by estimating $k$ object masks and $k$ object pivots from the motion network. This system requires placing a limitation on the number of objects, and doesn't yield better geometry estimation results than those from Zhou \etal~\cite{zhou2017unsupervised} or other systems~\cite{yang2018cvpr} which do not explicitly model moving objects. 

Optical flow estimation methods~\cite{horn1981determining} do consider dense 2D pixel matching, which is able to model both rigid motion because of camera movement and non-rigid motion induced by objects in the scene. 
Similar as in unsupervised depth learning, one may train a flow network in an unsupervised manner through view synthesis, as proposed recently by Jason \etal~\cite{jason2016back} and Ren \etal~\cite{ren2017unsupervised}. 
Although the learned flow network yields impressive results, these systems lack understanding of the underlining 3D geometry, yielding difficulties in regularization of the predictions, \eg~in the occluded regions. 

Some recent works~\cite{yin2018geonet, ranjan2018adversarial,zou2018dfnet} leverage the benefits from the two tasks. They either failed to consider rigid/non-rigid motion, occlusion regions, or didn't enforce consistency between the depth and optical flow. 

In this paper, we propose an effective unsupervised/self-supervised learning system by jointly considering the depth, camera pose and optical flow estimation via adaptive consistency. The motivation here is to better exploit the advantages of depth and optical flow. On non-occluded regions, 2D optical flow estimation is much easier and often more accurate than computing rigid flow via depth and motion. So it can be used for guiding depth and motion estimation. On the contrary, in occluded regions, there are no explicit cues for directly matching. We thus leverage depth and motion information to help optical flow estimation, as they are more reliable in this case. We call this adaptive consistency in contrast to the cross-task consistency proposed in DF-Net~\cite{zou2018dfnet}.
Our pipeline consider every pixel during the learning process, yielding significant performance boost on both geometry and motion estimation over previous SoTA methods (as illustrated in \figref{fig:example}. 

\begin{figure*}[t]
\centering
\includegraphics[width=\textwidth]{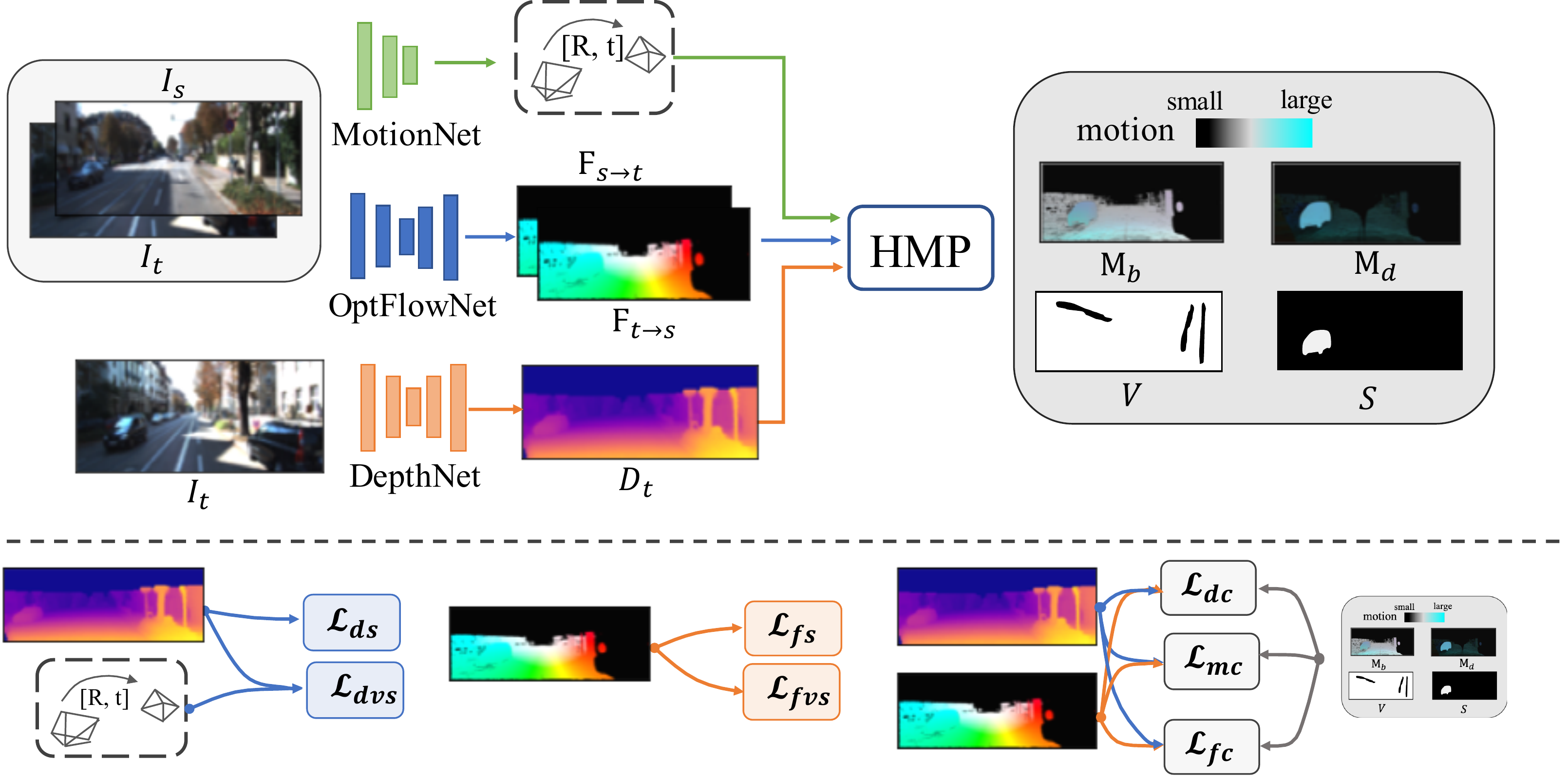}
\caption{The upper part of this figure shows the pipeline of our framework (EPC++). Given a a pair of consecutive frames, \ie target image $I_t$ and source image $I_{s}$, the OptFlowNet is used to predict optical flow $\ve{F}$ from $I_t$ to $I_{s}$. 
The MotionNet predicts their relative camera pose $\ve{T}_{t\rightarrow s}$. The DepthNet estimates the depth $\ve{D}_t$ from single frame. All three informations are fed into the Holistic 3D Motion Parser (HMP), which produce an segmentation mask for moving object $\ve{S}$, occlusion mask $\ve{V}$, 3D motion maps for rigid background $\ve{M}_b$ and dynamic objects $\ve{M}_d$. The bottom part of the figure shows how different loss terms are generated from geometrical cues. Details are shown in \secref{subsubsec:loss}}
\label{fig:pipeline}
\end{figure*}

We show the framework of EPC++ in \figref{fig:pipeline}.
Given two consecutive frames ($I_s$ and $I_t$), we estimate forward/backward flow maps ($\ve{F}_{t\rightarrow s}$, $\ve{F}_{s\rightarrow t}$), camera motion between the two frames ($\ve{T}_{t \rightarrow s}$) and corresponding depth maps ($\ve{D}_t, \ve{D}_s$).
The three types of information are fed into a holistic motion parser (HMP), where the visibility/non-occlusion mask ($\ve{V}$), the moving object segmentation mask ($\ve{S}$), the per-pixel 3D motions for rigid background ($\ve{M}_b$) and for moving object ($\ve{M}_d$) are recovered following geometrical rules and consistency. 
In principle, on non-occluded pixels, the values of $\ve{M}_d$ are encouraged to be close to zero in rigid regions, and to be large inside a moving object region, which yields the moving object mask. 
For pixels that are occluded, we use depth and camera motion to inpaint the optical flow, which shows more accurate results than using smoothness prior 
adopted by~\cite{ren2017unsupervised,wang2017occlusion}. 
We adopt the above adaptive consistency principles to guide the design of losses, and learning strategies for the networks. All the operations inside the HMP are easy to compute and differentiable. Therefore, the system can be trained end-to-end, which leverage the benefits of both depth estimation and optical flow prediction.

Last but not the least, recovering depth and object motion simultaneously from a monocular video, which is dependent on the given projective camera model~\cite{torresani2008nonrigid}, is an ill-posed problem. 
In particular, from the view point of a camera, a very close object moving with the camera is equivalent to a far object keeping relatively still, yielding scale confusion for depth estimation. Similar observations are also presented in \cite{godard2018digging}.  
Here, we address this issue by also incorporating stereo image pairs into our learning framework during training stage, resulting in a more robust system for depth, and optical flow estimation.

We conducted extensive experiments on the public KITTI 2015~\cite{geiger2012we}, Make3D~\cite{saxena2009make3d} and MPI-Sintel~\cite{Butler:ECCV:2012} dataset, and evaluated our results in multiple aspects including depth estimation, optical flow estimation, 3D scene flow estimation, camera motion and moving object segmentation. As elaborated in \secref{sec:exp}, EPC++ significantly outperforms other SoTA methods over all tasks. We will release the code of our paper upon its publication. 

In summary, the contributions of this paper lie in four aspects:
\begin{itemize}
    \item  We propose an effective unsupervised/self-supervised learning framework, EPC++, to jointly learn depth, camera motion, optical flow and moving object segmentation
    by leveraging the consistency across different tasks. 
    \item  We design a holistic motion parser (HMP) to decompose background foreground 3D motion with awareness of scene rigidity and visibility of each pixel.
    \item  We propose an adaptive learning strategy. It proves to be effective for training EPC++, which contains different coupled geometrical information.
    \item Comprehensive experiments over five tasks are conducted to validate each component in the proposed system. Results show that EPC++ achieves SoTA performance on all the tasks on KITTI datasets (driving scene), and also generalizes well to non-driving datasets such as Make3D and MPI-Sintel.
\end{itemize}

\section{Related Work}
\label{sec:related}

Estimating single view depth, predicting 3D motion and optical flow from images have long been central problems for computer vision. Here we summarize the most related works in various aspects without enumerating them all due to space limitation.

\noindent\textbf{Structure from motion and single view geometry.}
Geometric based methods estimate 3D from a given video with feature matching or patch matching, such as PatchMatch Stereo~\cite{bleyer2011patchmatch}, SfM~\cite{wu2011visualsfm}, SLAM~\cite{mur2015orb,engel2014lsd} and DTAM~\cite{NewcombeLD11}, and are effective and efficient in many cases. 
When there are dynamic motions in monocular videos, there is often scale-confusion for each non-rigid movement. Thus regularization through low-rank~\cite{dai2014simple}, orthographic camera~\cite{taylor2010non}, rigidity~\cite{kumar2017monocular} or fixed number of moving objects~\cite{kumar2016multi} are required in order to obtain an unique solution. 
However, those methods assumed the 2D matching are reliable, which can fail at where there is low texture, or drastic change of visual perspective \etc. More importantly, those methods can not be extended to single view reconstruction. 

Traditionally, estimating depth from a single view depends on hand-crafted features with specific and strong assumptions, such as computing vanishing point~\cite{HoiemEH07}, using assumptions of BRDF~\cite{prados2006shape,kong2015intrinsic}, or extracting the scene layout with major plane and box representations with Manhattan world~\cite{DBLP:conf/iccv/SchwingFPU13,DBLP:conf/3dim/SrajerSPP14} \etc. These methods typically only obtain sparse geometry representations. 

\noindent\textbf{Supervised depth estimation with CNN.}
Deep neural networks (DCN) developed in recent years provide stronger feature representation. Dense geometry, \ie~pixel-wise depth and normal maps, can be readily estimated from a single image~\cite{wang2015designing,eigen2015predicting,laina2016deeper,li2017two,cheng2018depth} and trained in an end-to-end manner. The learned CNN model show significant improvement compared to other methods which were based on hand-crafted features~\cite{karsch2014depth,ladicky2014pulling,zeisl2014discriminatively}. Others try to improve the estimation further by appending a conditional random field (CRF)~\cite{DBLP:conf/cvpr/WangSLCPY15,Liu_2015_CVPR,li2015depth,peng2016depth}. However, all these supervised methods require densely labeled ground truths, which are expensive to obtain in natural environments.
\noindent\textbf{Unsupervised single image depth estimation.}
Most recently, many CNN based methods are proposed to do single view geometry estimation with supervision from stereo images or videos, yielding impressive results. 
Some of them relies on stereo image pairs~\cite{xie2016deep3d,GargBR16,godard2016unsupervised}, \eg~ warping one image to another given known stereo baseline. 
Some others relies on monocular videos~\cite{zhou2017unsupervised,wang2017learning,li2017undeepvo,yang2018aaai,mahjourian2018unsupervised,yang2018cvpr} by incorporating 3D camera pose estimation from a motion network. However, as discussed in \secref{sec:introduction}, most of these models only consider a rigid scene, where moving objects were omitted. 
Vijayanarasimhan \etal~\cite{Vijayanarasimhan17} model rigid moving objects with $k$ motion masks, while their estimated depths were negatively effected by such an explicit rigid object assumption comparing to the one without object modeling~\cite{zhou2017unsupervised}. \chenxu{Casser \etal~\cite{casser2018depth} use Mask R-CNN~\cite{he2017mask} to generate possible moving regions and apply scale constraints on the moving objects. However, this can not be considered as a purely unsupervised or self-supervised method since it leverage supervisions from heterogeneous sources. }

Most theabove methods are solely based on photometric errors, \ie~$\|I_t(p_t) - \hat{I}_t(p_t)\|$, which use a Lambertian assumption, and are not robust in natural scenes with varing lighting conditions. To handle this problem, supervision signals based on local structural errors, such as local image gradient~\cite{yang2018aaai}, non-local smoothness~\cite{yang2018cvpr} and structural similarity (SSIM~\cite{wang2004image})~\cite{godard2016unsupervised,yin2018geonet} were proposed and yielded more robust matching and shows additional improvement on depth estimation. 
Most recently, Godard \etal~\cite{godard2018digging} further improved the results by jointly considering stereo and monocular images with updated neural network architectures.
In this work, we jointly consider the learning of optical flow network along with depth estimation and achieve SoTA performances for both depth and optical flow estimation. 

\noindent\textbf{Optical flow estimation.}
Similarly, there is a historical road map for optical flow estimation from traditional dense feature matching with local patterns, such as Patch matching~\cite{horn1981determining}, Piece-wise matching~\cite{black1996robust}, 
to supervised learning based on convolutional neural networks (CNNs), such as FlowNet~\cite{IMKDB17}, SPyNet~\cite{ranjan2017optical}, and PWCNet~\cite{sun2017pwc} \etc. These method produce significantly better performance due to deep hierarchical feature including larger while flexible context. Although these methods can be trained using synthesis datasets such as Flying Chairs~\cite{DFIB15} or Sintels~\cite{Butler:ECCV:2012}, they need high-quality labelled data of real-world scenes for good generalization, which is non-trivial to obtain~\cite{menze2015cvpr}.

The unsupervised learning of optical flow with a neural network is first introduced in \cite{ren2017unsupervised,jason2016back} by training CNNs with image synthesis and local flow smoothness.
Most recently, in \cite{wang2017occlusion,Meister:2018:UUL,wang2019unos}, the authors improve the results by explicitly computing the occlusion masks where photometric errors are omitted during the training, yielding more robust results. 
However, these works do not have 3D scene geometry understanding, \eg~depths and camera motion from the videos. In our case, we leverage such understanding and show a significant improvement over previous SoTA results. 

\noindent\textbf{3D Scene flow by joint depth and optical flow estimation.}
Estimating 3D scene flow~\cite{vedula1999three,vedula2005three} is a task of estimating per-pixel dense flow in 3D given a pair of images, which requires joint consideration of depths and optical flow of given consecutive frames. 
Traditional algorithms estimate depths from stereo images~\cite{menze2015cvpr,behl2017bounding}, or the given image pairs~\cite{kumar2017monocular} assuming rigid constraint, and trying to decompose the scene to piece-wise moving planes in order to finding correspondence with larger context~\cite{vogel2013piecewise,lv2016continuous}.
Most recently, Behl \etal~\cite{behl2017bounding} adopt semantic object instance segmentation and supervised stereo disparity from DispNet~\cite{mayer2016large} to solve large displacement of objects, yielding SoTA results on KITTI dataset. 

Most recently, works in unsupervised learning have begun to consider depths and optical flow together. Yin \etal~\cite{yin2018geonet} propose to estimate the residual flow in addition to the rigid flow, but the depth estimation did not benefit from the learning of optical flow. 
Ranjan \etal~\cite{ranjan2018adversarial} paste the optical flow from objects to the rigid flow from background and ego-motion to explain the whole scene in a competitive collaboration manner. However, rather than measuring 3D motion consistency, they divide the whole image with a selected threshold. DF-Net~\cite{zou2018dfnet} enforce consistency between rigid flow and optical flow but only in non-occluded and static regions.
In our case, we choose to model from the perspective of 3D scene flow, which is embedded in our unsupervised learning pipeline, yielding better results even with weaker backbone networks, \ie~VGG~\cite{simonyan2014very}, demonstrating the effectiveness of EPC++.

\noindent\textbf{Motion segmentation.}
Finally, since our algorithm decomposes static background and moving objects, our approach is also related to segmentation of moving objects from a given video. 
Current contemporary SoTA methods are dependent on supervision from human labels by adopting CNN image features~\cite{fragkiadaki2015learning,yoon2017pixel} or RNN temporal modeling~\cite{tokmakov2017learning}. 
For unsupervised video segmentation, saliency estimation based on 2D optical flow is often used to discover and track the objects~\cite{wang2018saliency,faktor2014video}, and long trajectories~\cite{brox2010object} of the moving objects based on optical flow need to be considered. 
However, these approaches commonly handle non-rigid objects within a relative static background, which is out of major scope of this paper. Most recently, Barnes \etal~\cite{barnes2018driven} show that explicitly modeling moving things with a 3D prior map can avoid visual odometry drifting. We also consider moving object segmentation, which is under an unsupervised setting with videos only. 

\section{Learning with Holistic 3D Motion Understanding}
\label{sec:approach}

As discussed in \secref{sec:introduction},
we obtain per-pixel 3D motion understanding by jointly modeling depth and optical flow, which is dependent on learning methods considering depth~\cite{zhou2017unsupervised} and optical flow~\cite{wang2017occlusion} independently. 

In the following, we will first elaborate on the geometry relationship between the two types of information, and then discuss the details about the how we leverage the rules of 3D geometry in EPC++ learning framework (\secref{subsec:scene_geometry}) through HMP.
Finally, we clarify all our loss functions and training strategies which consider both stereo and monocular images in training, with awareness of 3D motion dissected from HMP.

\subsection{Geometrical understanding with 3D motion}
\label{subsec:scene_geometry}
 Given two images, \ie~a target view image $I_t$ and a source view image $I_s$, suppose that $\ve{D}_t, \ve{D}_s$ are the depth maps of $I_t, I_s$, their relative camera
 transformation is $\ve{T}_{t\rightarrow s} = [\ve{R} | \ve{t}] \in \hua{S}\hua{E}(3)$ from $I_t$ to $I_s$,  and let optical flow from $I_t$ to $I_s$ be $\ve{F}_{t\rightarrow s}$.
 For one pixel $p_t$ in $I_t$, the corresponding pixel $p_s$ in $I_s$ can be found either through camera perspective projection  or with the optical flow, and they should be consistent. Formally, denote the corresponding pixel in source image $I_s$ found by optical flow as $p_{sf}$ and the matching pixel found by rigid transform as $p_{st}$, the computation can be written as,
\begin{align}
h(p_{st}) &= \pi(\ve{K}[\ve{T}_{t\rightarrow s}\ve{D}_t(p_t)\ve{K}^{-1}h(p_t) + \ve{M}^*_d(p_t)]), \nonumber \\
 p_{sf} &=  p_t + \ve{F}_{t\rightarrow s}(p_t),
\label{eqn:hmu}
\end{align}
where $\ve{D}_t(p_t)$ is the depth value of the target view at pixel $p_t$, and $\ve{K}$ is the camera intrinsic matrix, $h(p_t)$ is the homogeneous coordinate of $p_t$. $\pi(\ve{x})$ convert from the homogeneous coordinates to Cartesian coordinates, \ie~$\ve{x}/\ve{x}_{d}$ where $d$ is the vector dimension. Here, $d=3$ and the last element is the projected depth value at $p_s$ from $p_t$, which we represent it by $\hat{\ve{D}}_s(p_s)$.
$\ve{M}^*_d$ is the 3D motion of dynamic moving objects in the target camera coordinate.
In this way, every pixel in $I_t$ is explained geometrically.
Here, $p_s$ can be outside of the image $I_s$, or non-visible in $I_s$ when computing optical flow, which is also evaluated in optical flow estimation of KITTI dataset~\footnote{\url{http://www.cvlibs.net/datasets/kitti/eval_scene_flow.php?benchmark=flow}}.

Commonly, as proposed by previous works~\cite{zhou2017unsupervised,ren2017unsupervised,wang2017occlusion}, one may design CNN models for predicting $\ve{D}_t, T_{t\rightarrow s}, \ve{F}_{t\rightarrow s}$. After computed the corresponding $p_t$ and $p_s$, We can synthesize the target image $\hat{I}_t$ from the source image
and apply photometric loss as supervision:
\begin{align}
\hua{L}_p = \sum\nolimits_{p_t}\ve{V}(p_t)|I_t(p_t) - \hat{I}_t(p_t)|.
\label{eqn:photo}
\end{align}

Where $\ve{V}(p_t)$ is the visibility mask which is $1$ when $p_t$ is also visible in $I_s$, and $0$ if $p_t$ is occluded or falls out of view. Such models can be trained end-to-end.

By only considering $\ve{F}_{t\rightarrow s}$ in \equref{eqn:hmu}, and adding flow smoothness term yields unsupervised learning of optical flow~\cite{ren2017unsupervised,wang2017occlusion}.
On the other hand, dropping optical flow model, and assuming there is no dynamic motion in the scene, \ie~setting $\ve{M}^*_d = 0$ in \equref{eqn:hmu}, 
yields unsupervised learning of depths and motions~\cite{zhou2017unsupervised,yang2018cvpr}. 


In our case, to holistically model the 3D motion, we adopt CNN models for all the three components: optical flow, depths and motion. However, dynamic motion $\ve{M}_d$ and depths $\ve{D}_{s/t}$ are two conjugate pieces of information, where there always exists a motion pattern that can exactly compensate the error caused by inaccurate depth estimation. Considering matching $p_t$ and $p_s$ based on RGB could also be noisy, this yields an ill-posed problem with trivial solutions that prevent stable learning.
Therefore, we need to design effective learning strategies with strong regularization to provide effective supervision for all those networks, which we will describe later.

\noindent\textbf{Holistic 3D motion parser (HMP).}
In order to make the learning process feasible, we first need to distinguish between the motion from rigid background/camera motion and dynamic moving objects, regions of visible and occluded, where on visible rigid regions we can rely on structure-from-motion~\cite{zhou2017unsupervised} for training depths and on moving regions we can find 3D object motions.
As illustrated in \figref{fig:pipeline}, we handle this through a HMP that takes in the provided information from three networks, \ie~DepthNet, MotionNet and OptFlowNet, and outputs the desired dissected dense motion maps of background and moving things respectively.

Formally, given depths of both images $\ve{D}_s$ and $\ve{D}_t$, the learned forward/backward optical flow $\ve{F}_{t\rightarrow s / s\rightarrow t}$, and the relative camera pose $\ve{T}_{t\rightarrow s}$,  the motion induced by rigid background $\ve{M}_b$ and dynamic moving objects $\ve{M}_d$ from HMP are computed as,
\begin{align}
\ve{M}_b(p_t) &= \ve{T}_{t\rightarrow s}\phi(p_t|\ve{D}_t) - \phi(p_t|\ve{D}_t), \nonumber \\
\ve{M}_d(p_t) &= \ve{V}(p_t)[\phi(p_t+\ve{F}_{t\rightarrow s}(p_t)|\ve{D}_{s}) - \phi(p_t|\ve{D}_t) - \ve{M}_b(p_t)]  \nonumber \\
\ve{V}(p_t) &= \mathbbm{1}(\sum\nolimits_{p_s}(1-|p_t-(p_s + \ve{F}_{s\rightarrow t})|) > 0),\nonumber\\
\ve{S}(p_t) &= 1 - \exp\{-\alpha_s(\|\ve{M}_d(p_t)\|_2)\} 
\label{eqn:hmp}
\end{align}
where $\phi(p_t|\ve{D}_t) = \ve{D}_t(p_t)\ve{K}^{-1}h(p_t)$ is a back projection function from 2D to 3D space. Note here, different from $\ve{M}^*_d(p_t)$ in \equref{eqn:hmu}, $\ve{M}_d(p_t)$ only consider the dynamic per-pixel 3D motion at visible regions, which is easier to compute.
$\ve{V}$ is the visibility mask using the occlusion estimation from optical flow $\ve{F}_{s\rightarrow t}$ as presented in~\cite{wang2017occlusion}. We refer readers to their original paper for further details of the \yang{intuition and implementations}. $\ve{S}$ is a soft moving object mask, which indicates the confidence of a pixel that belongs to dynamic objects. 
$\alpha_s$ is a scaling hyper-parameter.

Here, we may further simplify the representation of $\ve{M}_d(p_t)$ by substituting $\ve{M}_b(p_t)$ in~\equref{eqn:hmp}, and put in the back projection function of $\phi()$ given the formula, \ie~
\begin{align}
    \ve{M}_d(p_t) &= \ve{V}(p_t)[\phi(p_t+\ve{F}_{t\rightarrow s}(p_t)|\ve{D}_{s}) -
                                 \ve{T}_{t\rightarrow s}\phi(p_t|\ve{D}_t)] \nonumber\\
                  &= \ve{V}(p_t)[\ve{D}_{s}(p_{sf})\ve{K}^{-1}h(p_{sf}) - \hat{\ve{D}}_{s}(p_{st})\ve{K}^{-1}h(p_{st})] \nonumber\\
                  &=  \ve{V}(p_t)\ve{K}[\ve{D}_{s}(p_{sf})h(p_{sf}) -                                           \hat{\ve{D}}_{s}(p_{st})h(p_{st})],
    \label{eqn:decompose}
\end{align}
Here, $\hat{\ve{D}}_{s}$ is the depth map of source image $I_s$ projected from the depth of target image $I_t$ as mentioned in \equref{eqn:hmu}. This will be useful for our loss design in \equref{loss:motion}.

After HMP, the rigid and dynamic 3D motions are disentangled from the whole 3D motion, and a moving object mask is estimated, where we could apply various supervision accordingly based on our structural error and regularization, which drives the joint learning of depth, motion and flow networks.

\subsection{Training the networks.}\label{subsec:training}
In this section, we will first introduce the networks for predicting and losses we designed for unsupervised learning.

\subsubsection{Network architectures.}
For depth prediction $\ve{D}$ and motion estimation between two consecutive frames $\ve{T}$ , we adopt the network architecture from Yang \etal~\cite{yang2018cvpr}, which depends on a VGG based encoder and double the input resolution of that used in Zhou \etal~\cite{zhou2017unsupervised}, \ie~$256 \times 832$, to acquire better ability in capturing image details.
In addition, for motion prediction, we drop the decoder for their explanability mask prediction since we can directly infer the occlusion mask and moving object masks through the HMP module to avoid error matching.

For optical flow prediction $\ve{F}$, rather than using FlowNet~\cite{IMKDB17} adopted in~\cite{wang2017occlusion}, we use a light-weighted network architecture, \ie~PWC-Net~\cite{sun2017pwc}, to learn a robust matching, which is almost 10$\times$ smaller than the FlowNet~\cite{IMKDB17}, while producing higher matching accuracy in our unsupervised setting.

We will describe the details of all these networks in our experimental section~\secref{sec:exp}.



\subsubsection{Training losses.} \label{subsubsec:loss}
After HMP~\equref{eqn:hmp}, the system generates various outputs, including: 1) depth map $\ve{D}$ from a single image $I$, 2) relative camera motion $\ve{T}$, 3) optical flow map $\ve{F}$, 4) rigid background 3D motion $\ve{M}_b$, 5) dynamic 3D motion $\ve{M}_d$, 6) visibility mask $\ve{V}$, and 7) moving object mask $\ve{S}$. Different loss terms are also used to effectively train corresponding networks as illustrated in pipeline shown in~\figref{fig:pipeline}.

\noindent\textbf{Rigid-aware structural matching.}
As discussed in \secref{sec:related}, photometric matching as proposed in~\equref{eqn:photo} for training flows and depths is not robust against illumination variations. In this work, in order to better capture local structures, we add additional matching cost from SSIM~\cite{wang2004image}, as applied in \cite{godard2016unsupervised}. In addition,
Formally, our matching cost can be written as, \ie~
\begin{align}
\scr{L}_{vs}(\ve{O})  &= \sum\nolimits_{p_t}\ve{V}_{*}(p_t, \ve{O})  s(I_t(p_t), \hat{I}_t(p_t)), \nonumber \\
\mbox{where,~~} & s(I_t(p_t), \hat{I_t}(p_t))= (1- \beta)  |I_t(p_t) - \hat{I_t}(p_t)| + \nonumber \\
&\mbox{~~~~~~~~} \beta  \frac{1 - \mbox{SSIM}(I_t(p_t), \hat{I}_t(p_t)) }{2}.
\label{eqn:str_matching}
\end{align}
Here, $\beta$ is a balancing hyper-parameter which is set to be $0.85$. $\ve{O}$ represents the type of input for obtaining the matching pixels, which could be $\ve{D}$ or $\ve{F}$ as introduced in~\equref{eqn:hmu}.
$\ve{V}_{*}$ indicates visibility mask dependent on the type of source image for synthesis. Specifically, for supervising depth, we needs to find rigid and non-occluded regions, and we let $\ve{V}_{d}(p_t, \ve{F}) = \ve{V}(p_t) (1 - \ve{S}(p_t))$. For supervising optical flow, we let $\ve{V}_{f}(p_t, \ve{F}) = \ve{V}(p_t)$ as~\equref{eqn:hmu}.
We denote view synthesis loss terms for depth and optical flow as $\scr{L}_{dvs}, \scr{L}_{fvs}$ respectively (as shown in \figref{fig:pipeline}).
Then, we may directly apply these losses to learn the flow, depth and motion networks.



\noindent\textbf{Edge-aware local smoothness.}
Although the structural loss alleviates the appearance confusion of view synthesis, the matching pattern is still a very local information. Therefore, smoothness is commonly adopted for further regularizing the local matching~\cite{revaud2015epicflow} to improve the results.
In our experiments, we tried two types of smoothness including edge-aware smoothness from image gradient proposed by Godard~\cite{godard2016unsupervised}, or smoothness with learned affinity similar to Yang~\etal~\cite{yang2018cvpr}.
We find that when using only photometric matching, the learned affinity provides significant improvements for final results over image gradient, but when adding structural loss (\equref{eqn:str_matching}), the improvements from learned affinity become marginal. From our perspective, this is mostly due to the robustness of the SSIM loss and the self-regularization from CNN. Therefore, in this work, for simplicity, we only use image gradient based edge-aware smoothness to regularize the learning of different networks, \chenxu{which is the same as used in ~\cite{godard2016unsupervised}.} Formally, the spatial smoothness loss can be written as,
\begin{align}
\scr{L}_{s}(\ve{O}) = \sum_{p_t}|\nabla^{2}\ve{O}(p_t)|e^{-\alpha_e|\nabla^2I(p_t)|},
\label{eqn:smoothness}
\end{align}
where $\ve{O}$ represents type of input. 
Here, we use $\scr{L}_{ds}$ and $\scr{L}_{fs}$ to denote the smoothness loss terms for depth and optical flow respectively. 

\noindent\textbf{Rigid-aware 3D motion consistency.}
Finally, we model the consistency between depths and flows in the rigid regions based on the outputs from our HMP.  
Specifically, we require $\ve{M}_d(p_t)$ to be small inside the rigid background regions, which can be calculated by $\ve{1} - \ve{S}$.
Formally, the loss functions can be written as,
\begin{align}
\hua{L}_{dmc} &= \sum\nolimits_{p_t} (1 - \ve{S}(p_t))|\ve{M}_d(p_t)|, \nonumber \\
\Leftrightarrow &\sum\nolimits_{p_t} (1 - \ve{S}(p_t)) \ve{V}(p_t)|\ve{D}_{s}(p_{sf})h(p_{sf}) -                                           \hat{\ve{D}}_{s}(p_{st})h(p_{st})| \nonumber \\
\label{loss:motion}
\end{align}
where $\ve{M}_d, \ve{S}(p_t)$ is defined in \equref{eqn:hmp}, and $\Leftrightarrow$ indicates equivalent in terms of optimization based on \equref{eqn:decompose}. 

In practice, we found the learning could be more stable by decomposing the 3D motion consistency to 2D flow consistency and depth consistency.
\chenxu{We hypothesize this is}
because the estimated depths at long distance can be much more noisy than the regions nearby (similar to the cases in supervised depth estimation~\cite{eigen2015predicting}), which induce losses difficult to minimize for the networks. Therefore, decomposing 3D motions to 2D motions and depths alleviates such difficulties.
Formally, we modify the optimization of original target to a new target by separately penalizing the difference over depths $\ve{D}_{s}(p_{sf})$ and flows $h(p_{sf})$ in $\hua{L}_{mc}$, \ie,
\begin{align}
    \hua{L}_{dmc} &= \hua{L}_{dc} + \hua{L}_{mc} \\
    \hua{L}_{dc} &= \sum\nolimits_{p_t} \ve{V}(p_t) (1 - \ve{S}(p_t)) (|\ve{D}_{s}(p_{sf}) -  \hat{\ve{D}}_{s}(p_{st})|  \nonumber\\
     \hua{L}_{mc} &= \sum\nolimits_{p_t} \ve{V}(p_t) (1 - \ve{S}(p_t))|p_{sf} - p_{st}|),
\label{eqn:dmc}
\end{align}
where $|\ve{D}_{s}(p_{sf}) - \hat{\ve{D}}_{s}(p_{st})|$ indicates the depth consistency, which is similar to the one used in~\cite{Vijayanarasimhan17}, and $|p_{sf} - p_{st}|$ indicates flow consistency inside rigid regions, which is similar to consistency check proposed in \cite{yin2018geonet}. However, we argue that these consistency are made to be more effective when combining with the masks estimated in our framework.
Here, we can see that the optima of $\hua{L}_{dmc}$ is also the optima for $\hua{L}_{mc}$, while the former is easier to optimize and is adopted in our training losses. 





\noindent\textbf{Flow motion consistency in occluded regions.}
Commonly, optical flow estimation on benchmarks, \eg~KITTI 2015~\cite{geiger2012we}, also requires flow estimation for pixels inside occlusion regions $\ve{V}$, which is not possible when solely using 2D pixel matching. Traditionally, researchers~\cite{wang2017occlusion, ren2017unsupervised} use local smoothness to ``inpaint'' those pixels from nearby estimated flows. Thanks to our 3D understanding, we can train those flows by requiring its geometrical consistency with our estimated depth and motion. Formally, the loss for 2D flow consistency is written as,
\begin{align}
\scr{L}_{fc} = \sum\nolimits_{p_t}(1 - \ve{V}(p_t))|p_{sf} - p_{st}|,
\end{align}
where $p_{sf},p_{st}$ are defined in \equref{eqn:decompose}. We use such a loss to drive the supervision of our OptFlowNet to predicting flows only at non-visible regions, and surprisingly, it also benefits the flows predicted at visible regions, which we think it is because well modeling of the occluded pixels helps regularization of training.

Nevertheless, one possible concern of our formula in 3D motion consistency is when the occluded part is from a non-rigid movement, \eg~a car moves behind another car. To handle this problem, it requires further dissecting object instance 3D motions, which we leave to our future work, and is beyond the scope of this paper.
In the datasets we experimented such as KITTI 2015, the major part of occlusion is from rigid background, which falls into our assumption. Specifically, we use the ground truth optical flow maps and moving masks from the validation images, and found $95\%$ of occluded pixels are in rigid background. 


\begin{figure*}
\centering
\includegraphics[width=\textwidth]{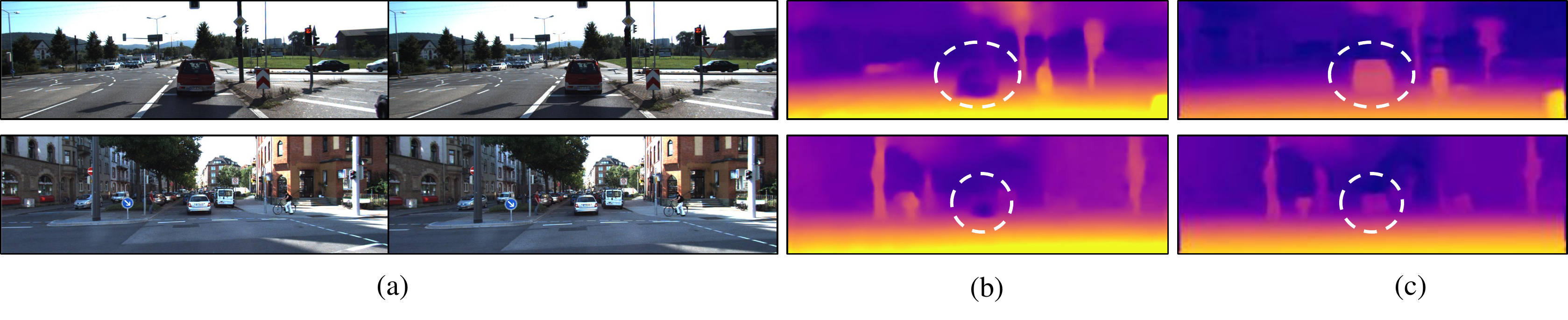}
\caption{Two examples of large depth confusion. Moving object in two consecutive frames (a) causes large depth value confusion for our system trained with monocular videos, as shown in (b). This issue can be resolved by incorporating stereo training samples into the system (c). }
\label{fig:infinity}
\end{figure*}

\noindent\textbf{Multi-scale penalization.}
Finally, in order to incorporate multi-scale context for training, following \cite{godard2016unsupervised} and \cite{zhou2017unsupervised}, we use four scales for the outputs of $\ve{D}$ and $\ve{F}$.
In summary, our loss functional for depths and optical flow supervision from a monocular video can be written as,
\begin{align}
\scr{L}_{mono}\! =\! \sum\nolimits_l &\lambda_{dvs}\scr{L}_{vs}^l(\ve{D_l}) + \lambda_{fvs}\scr{L}_{vs}^l(\ve{F_l}) \nonumber \\
&+ 2^l\lambda_{ds} \scr{L}_{s}^l(\ve{D_l}) + 2^l\lambda_{fs}\scr{L}_{s}^l(\ve{F_l}) \nonumber\\
&+ \lambda_{dc}\scr{L}^l_{dc} + \lambda_{mc}\scr{L}^l_{mc}+ \lambda_{fc}\scr{L}^l_{fc}
\label{eqn:full}
\end{align}
where $l$ indicates the level of image scale, and $l=0$ indicates the one with the highest resolution. $2^l$ is a weighting factor for balancing the losses between different scales. $\ve{\lambda} = [\lambda_{dvs}, \lambda_{fvs}, \lambda_{ds}, \lambda_{fs}, \lambda_{dc}, \lambda_{mc}, \lambda_{fc}]$ is the set of hyper-parameters balancing different losses, and we elaborate them in Alg.~\ref{alg:training}.

\subsubsection{Adaptive stage-wise learning strategy.}

In practice, we observe that jointly training all networks from scratch doesn't generates reasonable outputs. One possible reason is that many coupled geometrical cues (including parameters of all three MotionNet, OptFlowNet, DepthNet) are randomly initialized and thus generate very noise outputs at the beginning. Multiple noisy outputs (\eg ~$\ve{S},\ve{M}_b, \ve{M}_d$) make the learning difficult to converge.

Therefore, we adopt and alternative training by adaptively adjusting the hyper-parameters, \ie~$\ve{\lambda}$ and $\alpha_s$, as the training goes on, which switches on or off the learning of networks for more efficient convergence and also to serve as a better initialization for jointly learning of all networks.
Formally, we adopt a stage-wise learning strategy similar to~\cite{ranjan2018adversarial}, which trains the framework stage by stage and start the learning of later stages after previous stages are converged.
The learning algorithm is presented in Alg.~\ref{alg:training}.
First, we train DeptNet/MotionNet and OptFlowNet separately and there is no consistency enforced.
Then, after independent learning, we reset $\alpha_s$ to be a small constant $0.01$ to require the consistency over corresponding regions of the estimated depth, camera motion and optical flow.
In this stage, since the networks are continuously turning better, we alternatively optimize the depth and optical flow networks through iterative training. And we adaptively apply different masks ($\ve{S}, 1-\ve{S}, \ve{V}, 1-\ve{V}$,
\chenxu{where $1-\ve{S}$ is the inverted mask of $\ve{S}$, indicating the confidence score for the static regions and  $1-\ve{V}$ indicates the non-visibility score. }) in each iteration for better training of depth or optical flow networks.
In our experiments, the performance of all networks are saturated after two iterations in the alternative training stage, yielding SoTA performance for all the evaluated tasks, which we will elaborate in \secref{sec:exp}.

\begin{algorithm}
\DontPrintSemicolon
\SetAlgoLined
\KwResult{Trained \chenxu{networks} for predicting $\ve{D}$, $\ve{T}$, and $\ve{F}$}
\SetKwInOut{Input}{Input}\SetKwInOut{Output}{Output}
\Input{An unlabelled monocular video}
\BlankLine
 Define $\ve{\lambda} = [\lambda_{dvs}, \lambda_{fvs}, \lambda_{ds}, \lambda_{fs}, \lambda_{dc}, \lambda_{mc}, \lambda_{fc}]$ as loss balancing parameters in~\equref{eqn:full}.\;
 Set $\alpha_s = 0$ for moving mask computation in~\equref{eqn:hmp}.\;
 \textbullet Train Depth and Motion networks till convergence with $\ve{\lambda} = [1, 0, 1, 0, 0, 0, 0]$.\;
 \textbullet Train Optical flow network till convergence with $\ve{\lambda} = [0, 1, 0, 1, 0, 0, 0]$.\;
 \textbullet Re-set $\alpha_s = 0.01$ for moving mask computation. \;
 \While{}{
 \textbullet Train Depth and Motion networks guided by optical flow till convergence with $\ve{\lambda} = [1, 0, 1, 0, 0.05, 0.25, 0]$.\\
 \textbullet Train Optical flow network guided by depth flow till convergence with $\ve{\lambda} = [0, 1, 0, 1, 0, 0, 0.005]$.
 }
 \caption{Training EPC++ with monocular videos over the KITTI 2015 dataset. At each step we train the network until convergence (details in \secref{subsec:impl_detail}).}
 \label{alg:training}
\end{algorithm}

\subsection{Stereo to solve motion confusion.}
\label{sec:motion_confusion}
As discussed in the introduction part (\secref{sec:introduction}), the reconstruction of moving objects in monocular video has projective confusion, which is illustrated in \figref{fig:infinity}. The depth map in (b) is an example predicted with our algorithm trained with monocular samples, where the car in the front is running at the same speed and the region is estimated to be far. This is because when the depth value is estimated large, the car will stay at the same place in the warped image, yielding small photometric errors during training, as also observed in \cite{godard2018digging}.
Obviously, the losses of motion or smoothness~\equref{eqn:full} does not solve this issue. Therefore, we have added stereo images (which are captured at the same time but from different view points) into learning the depth network to avoid such confusion jointly with monocular videos. As shown in \figref{fig:infinity} (c), the framework trained with stereo pairs correctly figures out the depth of the moving object regions.

Formally, when corresponding stereo image $I_{c}$ is additionally available for the target image $I_{t}$, we treat $I_{c}$ as another source image, similar to $I_{s}$, but with known camera pose $\ve{T}_{t\rightarrow c}$. In this case, since there is no motion factor (stereo pairs are simultaneously captured), we adopt the same loss of $\hua{L}_{s}$ and $\hua{L}_{vs}$ taking $I_{c}, I_{t}$ as inputs for supervising the depth network. Formally, the total loss for DepthNet when having stereo images is,
\begin{align}
\scr{L}_{mono_stereo}& =\scr{L}_{mono}  \nonumber\\
        &+ \sum\nolimits_l\{\lambda_{cvs}\scr{L}_{vs}^l(I_c) + \lambda_{cs}\scr{L}_{s}^l(I_c) \}.
\label{eqn:stereo_full}
\end{align}
where $\scr{L}_{}(I_c)$ and $\scr{L}_{bi-vs}(I_c)$ indicate the corresponding losses with a visibility mask  computed using stereo image $I_c$. Here, we update steps of learning depth and motion networks in Alg.~\ref{alg:training} by adding the loss from stereo pair with $\lambda_{cvs} = 4$ and $\lambda_{cs} = 10$.

\section{Experiments}
\label{sec:exp}

In this section, we firstly describe the datasets and evaluation metrics used in our experiments, and then present comprehensive evaluation of EPC++ on different tasks.

\subsection{Implementation details}
\label{subsec:impl_detail}

EPC++ consists of three sub-networks: DepthNet, OptFlowNet and MotionNet as described in \secref{sec:approach}. 
Our HMP module has no learnable parameters, thus does not increase the model size. In the following, we clarify the network architectures, corresponding training procedure and setting of hyper-parameters.

\begin{figure}
\centering
\includegraphics[width=0.5\textwidth]{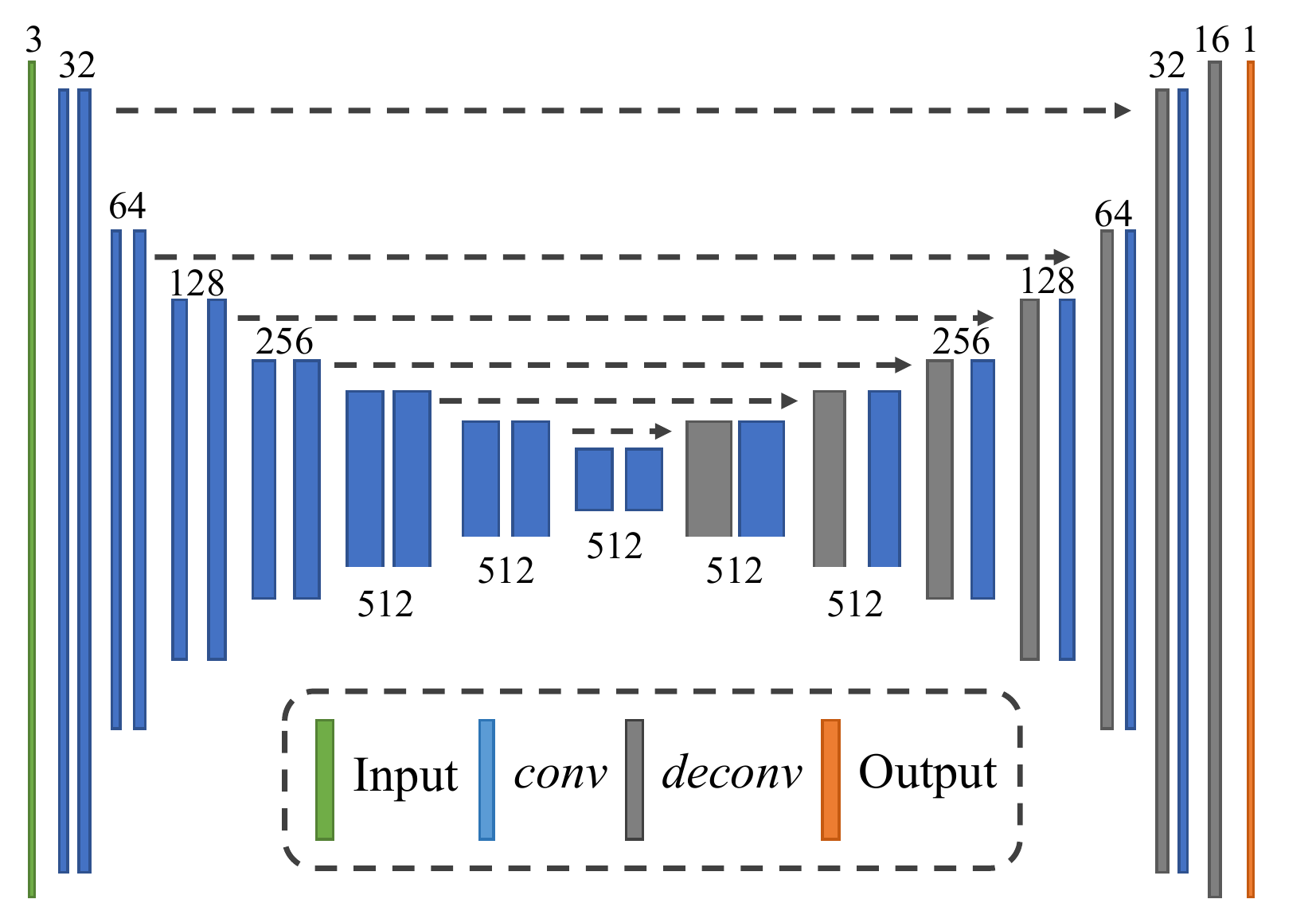}
\caption{The architecture of DepthNet. \yang{Each rectangle represents one certain layer as color coded in the legend. Number on top of the rectangles indicates the channel size of each layer (rectangle), e.g. 32 in the left indicates both convolutional layers have 32 channels.}}
\label{fig:dispnet_arc}
\end{figure}

\noindent\textbf{Training DepthNet.}
We modify the DispNet \cite{mayer2016large} architecture for DepthNet as illustrated in  \figref{fig:dispnet_arc}.
Our dispNet is based on an encoder-decoder design with skip connections and multi-scale side outputs. The encoder consists of 14 convolutional layers with kernel size of 3 except for the first 4 \textit{conv} layers with kernel size of {7, 7, 5, 5} respectively.
The decoder has symmetrical architecture as the encoder, consisting of 7 \textit{conv} layers and 7 \textit{deconv} layers. To capture more details, both input and output scales of the DepthNet are set as $256 \times 832$, which is twice as large as used in \cite{zhou2017unsupervised}. 

All \textit{conv} layers are followed by ReLU activation except the output layer where we apply a sigmoid activation to constrain the depth prediction within a reasonable range. In practice, output disparity range is constrained within 0-0.3.
Batch normalization (BN) \cite{ioffe2015batch} is performed on all \textit{conv} layers when training with stereo videos, and is dropped when training with monocular videos for better stability and performance.
This is because BN helps to reduce the scale variation between monocular and stereo images.
Last, for stereo training, following~\cite{godard2016unsupervised}, we ask the DepthNet to output the disparity maps of both the left and the right images for computing left-right consistency.

For training, the Adam optimizer \cite{kingma2014adam} is applied with $\beta_1=0.9$, $\beta_2=0.999$, learning rate of $2\times 10^{-4}$ and batch size of $4$.
In independent training stage, the loss balance for DepthNet are set as $\lambda_{dvs}=1.0$, $\lambda_{ds}=1.0$ in \equref{eqn:full} as shown in Alg.~\ref{alg:training} respectively. The model is trained for 180K iterations. In alternative training stage, we decrease the learning rate to $2\times 10^{-5}$ and train the network for about 10K iterations.

For training using stereo videos, the DepthNet is trained with $\lambda_{ds}=1.0$, $\lambda_{dvs}=2.5$ in \equref{eqn:stereo_full}, for 145,000 iterations in independent training stage and hyper-parameters and training iterations are set to be the same.

\noindent\textbf{Training MotionNet.}
The MotionNet architecture is the same as the Pose CNN in \cite{zhou2017unsupervised} which outputs a  6-dimensional vector for camera motion. 
The training process is identical to the DepthNet since they need to be trained together.

\noindent\textbf{Training OptFlowNet.}
We use PWC-Net \cite{sun2017pwc} as our OptFlowNet. PWC-Net is based on an encoder-decoder design with intermediate layers warping CNN features for reconstruction. 
The batch size is set as 4 and we set $\lambda_{fvs}=1.0$, $\lambda_{fs}=1$.
During independent training stage, the network is optimized with Adam optimizer \cite{kingma2014adam} with $\beta_1=0.9$, $\beta_2=0.999$, learning rate of $2\times 10^{-4}$ for 100,000 iterations from scratch. During adaptive training stages, the learning rate is decreased to $2\times 10^{-5}$, and the network is trained with about 30K iterations.

\noindent\textbf{Hyper parameters.}  We set $\beta=0.85$ in  \equref{eqn:str_matching} and $\alpha_e=10$ in \equref{eqn:smoothness} following~\cite{wang2017occlusion}.
For $\alpha_s$ in \equref{eqn:hmp},  we validate it through a validation set using depth metrics within a set of $\{0.01, 0.05, 0.1, 0.5\}$. In general, there is no significant difference for the final performance, \ie ranging from 0.144 to 0.146 for the relative absolute error. Here we pick the best one, \ie $\alpha_s=0.01$ as shown in Alg~\ref{alg:training}. For $\lambda_{dc}$, $\lambda_{mc}$ and $\lambda_{fc}$ for depth-flow consistency, we use similar strategy as tuning $\alpha_s$, and set to be $0.05$, $0.25$ and $0.005$ correspondingly during the adaptive training stage for monocular setting as illustrated in Alg~\ref{alg:training}. As for stereo setting, we set $\lambda_{fc}$ to be $0.02$ with other hyper parameters remain the same.

EPC++ has 38.3M parameters in total, in which DepthNet and MotionNet have 33.2M parameters and OptFlowNet has 5.1M parameters.

\subsection{Datasets and metrics}
Extensive experiments were conducted on five tasks to validate the effectiveness of EPC++ in different aspects. These tasks include: depth estimation, optical flow estimation, 3D scene flow estimation, odometry and moving object segmentation. All the results are evaluated on the KITTI dataset using the corresponding standard metrics commonly used by other SoTA methods~\cite{yang2018cvpr,wang2017occlusion,yin2018geonet,ranjan2018adversarial}.

\noindent\textbf{KITTI 2015.}
The KITTI 2015 dataset provides videos in 200 street scenes captured by stereo RGB cameras, with sparse depth ground truths captured by Velodyne laser scanner. 2D flow and 3D scene flow ground truth are generated from the ICP registration of the point cloud projection. The moving object mask is provided as a binary map to distinguish between static background and moving foreground in flow evaluation. During training, 156 stereo videos that exclude test and validation scenes are used. The monocular training sequences are constructed with three consecutive frames; left and right views are processed independently. This leads to 40,250 monocular training sequences. Stereo training pairs are constructed with left and right frame pairs, resulting in a total of 22,000 training samples.

\begin{table}[!htbp]
\centering
\fontsize{8}{10}\selectfont
\def\arraystretch{1.5}
\caption{Evaluation metrics for our tasks. From top row to bottom row: depth, optical flow, odometry, scene flow and segmentation.}
\setlength{\tabcolsep}{2pt}
\begin{tabular}{l|l}
\specialrule{.15em}{.1em}{.1em}
Abs Rel: $\!\frac{1}{|D|}\!\sum_{d'\in D}\!|d^*\!\!\!-\!\!d'|/d^*$       & Sq Rel: $\frac{1}{|D|}\!\sum_{d'\in D}\!||d^*\!\!\!-\!\!d'||^2\!/d^*$                \\
RMSE: $\!\sqrt{\!\frac{1}{|D|}\!\sum_{d'\!\in\! D}||d^*\!\!\!-\!\!d'||^2}$    & RMSE log: $\!\sqrt{\!\!\frac{1}{|D|}\!\!\sum_{d'\!\in\! D}\!\!||\!\log\! d^*\!\!\!-\!\!\log\! d'||^2\!}\!$ \\
$\delta_t$: $\%$ of $d \in D$ $max(\frac{d^*}{d}, \frac{d}{d^*})<t$ \\
\hline
EPE: $\frac{1}{|F|}\sqrt{\sum_{f'\in F}\|f^*-f'\|^2}$ & F1: err $>$ 3px and err $>|f^*|\times 5\%$ \\
\hline
\multicolumn{2}{l}{ATE, $t_{err}$: $\frac{1}{|T|}\sqrt{\sum_{t'\in T}\|t^*-t'\|^2}$}  \\
\multicolumn{2}{l}{$r_{err}$: $\frac{1}{|T|}\sum [\arccos\frac{Tr(R')-1}{2}-\arccos\frac{Tr(R)-1}{2}]$}
 \\ \hline
D1, D2:  $\!\frac{1}{|D|}\!\sum_{d'\in D}\!|d^*\!\!\!-\!\!d'|$        & FL: $\!\frac{1}{|F|}\!\sum_{f'\in F}\!|f^*\!\!\!-\!\!f'|$         \\
\hline
pixel acc: $\!\frac{\sum_i n_{ii}}{\sum_i t_i}$ & mean acc:    $\!\frac{1}{n_{cl}}\sum_i\frac{n_{ii}}{t_i}$           \\
mean IoU: $\!\frac{1}{n_{cl}}\sum_i\frac{n_{ii}}{t_i+\sum_j n_{ji}-n_{ii}}$ & f.w. IoU: $\!\frac{1}{\sum_k t_k}\sum_i\frac{t_in_{ii}}{t_i+\sum_j n_{ji}-n_{ii}}$ \\
\hline
\end{tabular}
\label{metrics}
\end{table}

For \textit{depth} evaluation, we chose the Eigen split \cite{eigen2014depth} for experiments to compare with more baseline methods. The Eigen test split consists of 697 images, where the depth ground truth is obtained by projecting the Velodyne laser scanned points into the image plane. To evaluate at input image resolution, we re-scale the depth predictions by bilinear interpolation. The sequence length is set to be 3 during training.
For \textit{optical flow} evaluation, we report performance numbers on both training and test splits of KITTI 2012 and KITTI 2015 datasets and compare with other unsupervised methods. Both training and test set contain 200 image pairs. Ground truth optical flow for training split is provided and the ground truth for test split is withheld on the official evaluation server.

For \textit{scene flow} and \textit{segmentation} evaluation, we evaluate on the KITTI 2015 training split, containing 200 image pairs. The scene flow ground truth is publicly available and the moving object ground truth is only provided for this split.
KITTI 2015 dataset also provides an \textit{odometry} data split, consisting of 9 training sequences (Seq. 00-08) and 2 test sequences (Seq. 09, Seq. 10). On average, there are 2,200 frames in one training sequence, resulting in over 20,000 training samples. Seq.09 and Seq. 10 contain about 1,200 and 1,500 frames respectively.


\begin{table*}[!ht]
\centering
\caption{Single view depth estimation results on the Eigen test split. Methods trained with monocular samples are presented in the upper part and those also taking stereo pairs for training are presented in the bottom part. All results are generated by models trained on KITTI data only unless specially noted. Details are in \secref{subsec:depth_eval}.}
\label{tbl:sota}
\fontsize{8}{9.5}\selectfont
\setlength{\tabcolsep}{5pt}
\bgroup
\def\arraystretch{1.2}
\begin{tabular}{lcccccccccc}
\specialrule{.2em}{.1em}{.1em}
\multirow{2}{*}{Method}      & \multirow{2}{*}{Stereo} &  & \multicolumn{4}{c}{Lower the better}                     &  & \multicolumn{3}{c}{Higher the better}                                                                     \\ \cline{4-7} \cline{9-11}
                                                                                 &  &                           & Abs Rel        & Sq Rel         & RMSE  & RMSE log       &  & $\delta \textless 1.25$ & $\delta \textless 1.25^2$ & $\delta \textless 1.25^3$ \\ \hline
\multicolumn{1}{l|}{Train mean}       &                         &  & 0.403          & 5.530          & 8.709 & \multicolumn{1}{l|}{0.403}          &  & 0.593                 & 0.776                                   & 0.878                                   \\
\multicolumn{1}{l|}{SfMLearner ~\cite{zhou2017unsupervised}}                            &                          &  & 0.208          & 1.768          & 6.856 & \multicolumn{1}{l|}{0.283}          &  & 0.678                 & 0.885                                   & 0.957                                   \\
\multicolumn{1}{l|}{LEGO\cite{yang2018cvpr}}        &                     &      & 0.162          & 1.352          & 6.276 & \multicolumn{1}{l|}{0.252}          &  & 0.783                 & 0.921                                   & 0.969\\
\multicolumn{1}{l|}{Mahjourian \etal \cite{mahjourian2018unsupervised}}     & &  & 0.163          & 1.240          & 6.220 & \multicolumn{1}{l|}{0.250}          &  & 0.762      & 0.916                                   & 0.968 \\
\multicolumn{1}{l|}{DDVO \cite{wang2018learning}}     & &  & 0.151          & 1.257          & 5.583 & \multicolumn{1}{l|}{0.228}          &  & 0.810      & 0.936                                   & 0.974 \\
\multicolumn{1}{l|}{GeoNet-ResNet(update)\cite{yin2018geonet}}                       &  &                          & 0.149&	1.060&	5.567& 	\multicolumn{1}{l|}{0.226}& &	0.796	&0.935	&0.975 \\
\multicolumn{1}{l|}{Competitive-Collaboration~\cite{ranjan2018adversarial}}                       &  &                          & 0.148&	1.149&	5.464& 	\multicolumn{1}{l|}{0.226}& &	0.815	&0.935	&0.973 \\
\multicolumn{1}{l|}{DF-Net \cite{zou2018dfnet} (ResNet-50)}                     &  &                         &  0.145 &	 1.290	& 5.612	& \multicolumn{1}{l|}{0.219} & &	0.811&0.939	&\textbf{0.977} \\ \hline
\multicolumn{1}{l|}{EPC++ (mono depth only)}  &     &  &  0.151 & 1.448 & 5.927 & \multicolumn{1}{l|}{0.233} & & 0.809 & 0.933 & 0.971 \\
\multicolumn{1}{l|}{EPC++ (mono depth consist)}                                &  &    &  0.146 &  1.065 &  5.405 &  \multicolumn{1}{l|}{0.220} & &0.812 & 0.939 & 0.975 \\ \hline
\multicolumn{1}{l|}{\chenxu{EPC++ (mono joint w/ flow)}}                                &    &  &  0.156 & 1.075 & 5.711 & \multicolumn{1}{l|}{0.229} & & 0.783 & 0.931 & 0.974\\
\multicolumn{1}{l|}{EPC++ (mono flow consist)}                                 &    &  &  0.148 & 1.034 & 5.546 & \multicolumn{1}{l|}{0.223} & & 0.802 & 0.938 & 0.975\\
\multicolumn{1}{l|}{EPC++ (mono vis flow consist)}                                 &    &  & 0.144 & 1.042 & 5.358 & \multicolumn{1}{l|}{0.218} & & 0.813 & 0.941 & 0.976 \\
\multicolumn{1}{l|}{EPC++ (mono)}      &               &  & \textbf{0.141} & \textbf{1.029} & \textbf{5.350} & \multicolumn{1}{l|}{\textbf{0.216}} &  & \textbf{0.816}        & \textbf{0.941}                          & 0.976                          \\  \hline
\multicolumn{1}{l|}{UnDeepVO\cite{li2017undeepvo}}     &         \checkmark &               & 0.183          & 1.730          & 6.570 & \multicolumn{1}{l|}{0.268}          &  & -                     & -                                       & -                                       \\
\multicolumn{1}{l|}{Godard \etal\cite{godard2016unsupervised}}        & \checkmark              &  & 0.148          & 1.344          & 5.927 & \multicolumn{1}{l|}{0.247}          &  & 0.803                 & 0.922                                   & 0.964    \\
\multicolumn{1}{l|}{EPC \cite{yang2018every}}  &\checkmark     &  &  \textbf{0.127} & 1.239 & 6.247 & \multicolumn{1}{l|}{0.214} & & \textbf{0.847} & 0.926 & 0.969 \\
\multicolumn{1}{l|}{EPC++ (stereo depth only)}     &         \checkmark &               & 0.141          & 1.224          & 5.548 & \multicolumn{1}{l|}{0.229}          &   & 0.811                     & 0.934                           & 0.972                                       \\
\multicolumn{1}{l|}{EPC++ (stereo depth consist)}     &         \checkmark &               & 0.134          & 1.063          & 5.353 & \multicolumn{1}{l|}{0.218}          &  & 0.826                     & 0.941                         & 0.975                                       \\
\multicolumn{1}{l|}{EPC++ (stereo)}     &         \checkmark &               & \textbf{0.127}          & \textbf{0.936}          & \textbf{5.008} & \multicolumn{1}{l|}{\textbf{0.209}}          &  & 0.841                     & \textbf{0.946}                                       & \textbf{0.979}                                       \\\hline
\end{tabular}
\egroup
\end{table*}

\noindent\textbf{Make3D.} Make3D dataset~\cite{saxena2009make3d} contains no videos but 534 monocular image and depth ground truth pairs. Unstructured outdoor scenes, including bush, trees, residential buildings, etc. are captured in this dataset. Same as in~\cite{zhou2017unsupervised,wang2018learning}, the evaluation is performed on the test set of 134
images.
\noindent\textbf{MPI-Sintel. } MPI-Sintel dataset ~\cite{Butler:ECCV:2012} is obtained from
an animated movie which pays special attention to realistic image effects. It contains multiple sequences including large/rapid motions. We use the ``final'' pass of the data to train and test our model, which consists of 1,000 image pairs.
\noindent\textbf{Metrics.} The existing metrics of depth, optical flow, odometry, segmentation and scene flow were used for evaluation, as in previous methods \cite{eigen2014depth,menze2015cvpr,long2015fully}. For depth and odometry evaluation, we adopt the code from \cite{zhou2017unsupervised}. For optical flow and scene flow evaluation, we use the official toolkit provided by \cite{menze2015cvpr}.

For foreground segmentation evaluation, we use the overall/per-class pixel accuracy and mean/frequency weighted (f.w.) IOU for binary segmentation.
The definition of each metric used in our evaluation is specified in Tab. \ref{metrics}, in which, $x^*$ and $x'$ are ground truth and estimated results ($x \in \{d, f, t\}$). $n_{ij}$ is the number of pixels of class $i$ segmented into class $j$. $t_j$ is the total number of pixels in class $h$. $n_{cl}$ is the total number of classes.


\subsection{Depth evaluation}
\label{subsec:depth_eval}

\noindent\textbf{Experiment setup.}
The depth experiments are conducted on KITTI Eigen split~\cite{eigen2014depth} to evaluate the performance of EPC++ and its variants. 
The depth ground truths are sparse maps as they come from the projected Velodyne Lidar points.
Only pixels with ground truth depth values (valid Velodyne projected points) are evaluated. For monocular model, following~\cite{zhou2017unsupervised}, we scale the predicted depth to match the median with the groundtruth. For stereo model, we use the given intrinsic and baseline to compute the depth from estimated disparity, which is the same as~\cite{godard2016unsupervised}
The following evaluations are performed to present the depth performances: (1) ablation study of our approach and (2) depth performance comparison with the SoTA methods.

\noindent\textbf{Ablation study.}
We explore the effectiveness of each component of EPC++ as presented in Tab. \ref{tbl:sota}. Several variant results are generated for evaluation, including:

\noindent(1) EPC++ (mono depth only): DepthNet trained with view synthesis and smoothness loss ($\hua{L}_{dvs}+\hua{L}_{ds}$) on monocular sequences without visibility masks, which is already better than many SoTA methods. This is majorly due to the SSIM introduced in structural matching~\cite{godard2016unsupervised}, our modified DepthNet with higher resolution inputs~\cite{yang2018cvpr}, and depth normalization as introduced in~\cite{wang2018learning}.

\noindent(2) EPC++ (mono depth consist): Fine-tune the trained DepthNet with a depth consistency term as formulated with  $\hua{L}_{dc}=|\ve{D}_{s}(p_{sf}) -  \hat{\ve{D}}_{s}(p_{st})|$ term, which is a part of \equref{eqn:dmc}; we show it benefits the depth learning.

\noindent(3) \chenxu{EPC++ (mono joint w/ flow consist): Fine-tune the whole system with depth/flow consistency (see \equref{eqn:dmc}). As we can see, joint training results in worse depth performance and also worse flow performance (see optical flow evaluation). }

\noindent(4) EPC++ (mono flow consist): DepthNet trained by adding flow consistency in \equref{eqn:dmc}, where we drop the visibility mask. We can see that the performance is worse than adding depth consistency alone since flow at non-visible parts harms the matching.

\noindent(5) EPC++ (mono vis flow consist): DepthNet trained with depth and flow consistency as in \equref{eqn:dmc}, but add the computation of visibility mask $\ve{V}$; this further improves the results.

\noindent(6) EPC++ (mono): Final results from DepthNet with two iterations of adaptive depth-flow consistency training, yielding the best performance among all monocular trained methods.

We also explore the use of stereo training samples in our framework, and report performances of two variants: (6) EPC (stereo depth only): DepthNet trained on stereo pairs with only $\hua{L}_{dvs}+\hua{L}_{ds}$.

\noindent(7) EPC++ (stereo depth consist): DepthNet trained on stereo pairs with depth consistency.

\noindent(8) EPC++ (stereo): Our full model trained with stereo samples.

It is notable that for monocular training, the left and right view frames are considered independently and thus the frameworks trained with either monocular or stereo samples leverage the same amount training data. As shown in Tab. \ref{tbl:sota}, our approach (EPC++) trained with both stereo and sequential samples shows large performance boost over using only one type of training samples, proving the effectiveness of incorporating stereo into the training. With fine-tuning from HMP, comparing results of EPC++ (stereo) and EPC++ (stereo depth consist), the performance is further improved.


\begin{figure*}
\centering
\includegraphics[width=1.0\textwidth]{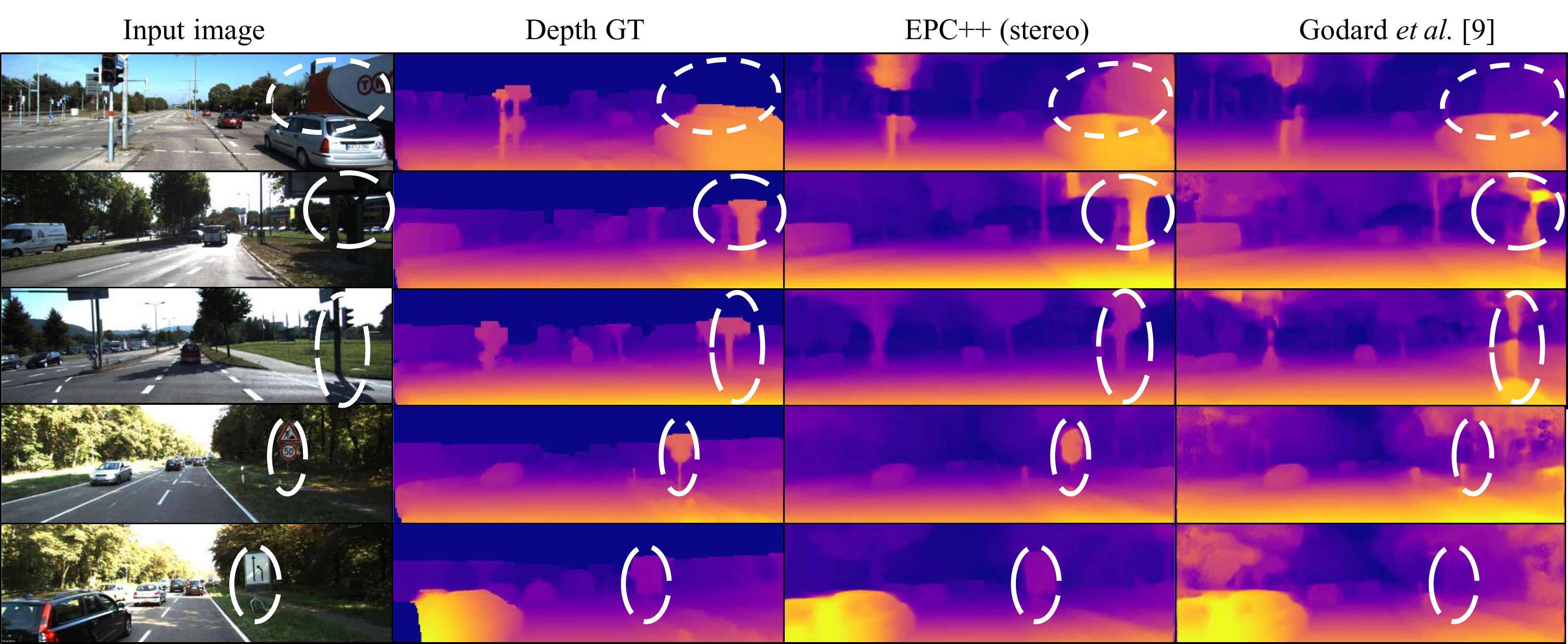}
\caption{Visual comparison between Godard \etal \protect\cite{godard2016unsupervised} and EPC++ (stereo) results on KITTI frames. The depth ground truths are interpolated and all images are reshaped for better visualization. For depths, our results have preserved the details of objects noticeably better (as in white circles).}
\label{fig:stereo_examples}
\end{figure*}

\begin{figure*}
\centering
\includegraphics[width=1.0\textwidth]{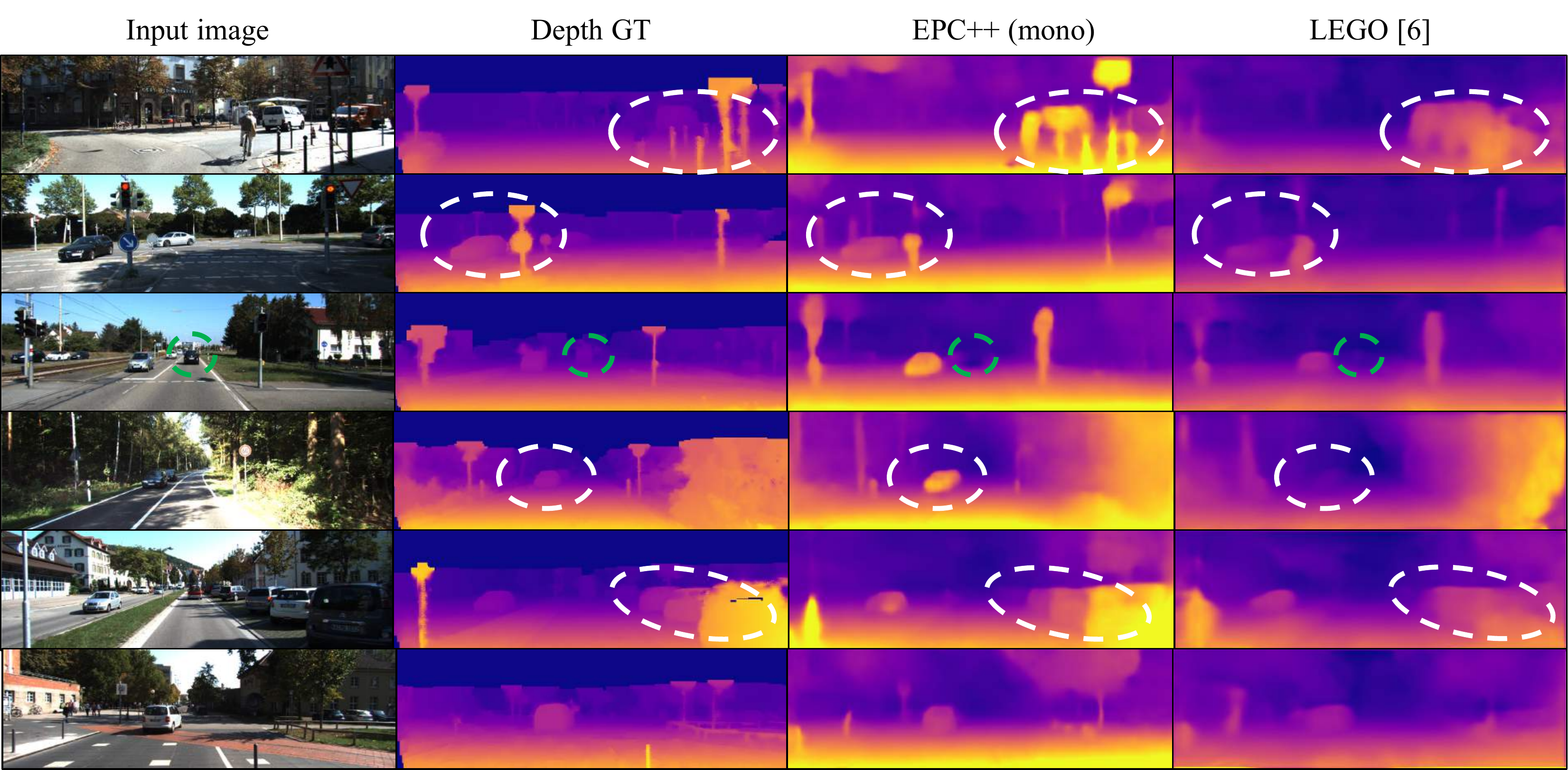}
\caption{Visual comparison between LEGO \protect\cite{yang2018cvpr} and EPC++ (mono) results on KITTI test frames. Thanks to the extra supervision from optical flow, our monocular results preserve the details of the occluded/de-occluded regions better, \textit{e.g.} the structure of thin poles. Please note the ``large depth value confusion'' still happens for both monocular based methods (green circle).}
\label{fig:mono_examples}
\end{figure*}

\begin{figure*}
\centering
\includegraphics[width=1.0\textwidth]{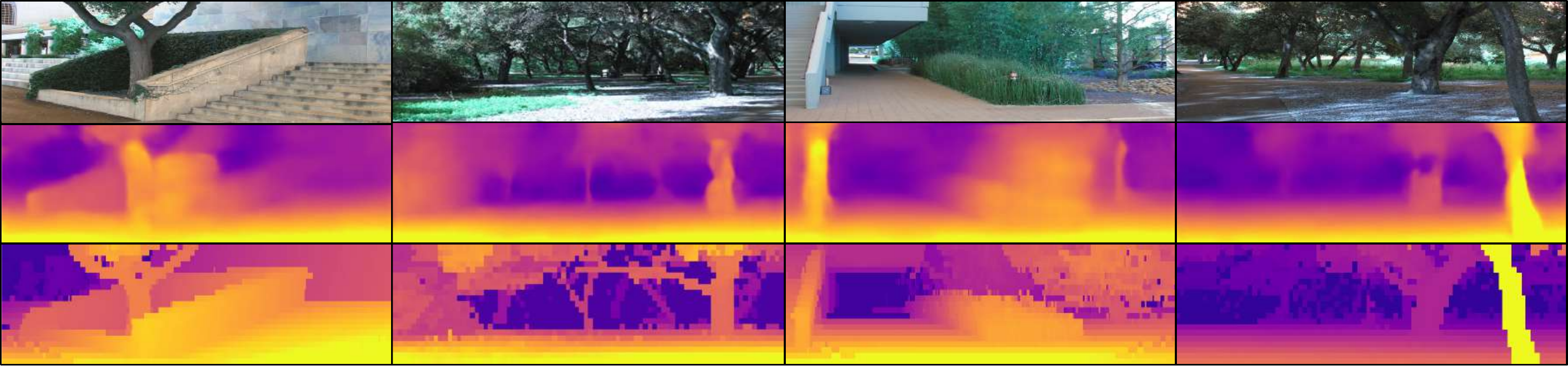}
\caption{Qualitative depth estimation results on Make3D. The results are generated by applying EPC++, which is trained on KITTI dataset, on Make3D test images. From top to bottom: input test image, our depth estimation result, depth ground truth.}
\label{fig:make3d_depth}
\end{figure*}

\noindent\textbf{Comparison with state-of-the-art.}
Following the tradition of other methods \cite{eigen2014depth,zhou2017unsupervised,godard2016unsupervised}, 
the same crop as in \cite{eigen2014depth} is applied during evaluation on Eigen split. We conducted a comprehensive comparison with SoTA methods that take both monocular and stereo samples for training.

Tab. \ref{tbl:sota} shows the comparison of EPC++ and recent SoTA methods. Our approach outperforms current SoTA unsupervised methods \cite{zhou2017unsupervised,kuznietsov2017semi,yang2018aaai,yang2018cvpr,godard2016unsupervised} on all metrics by a large margin. It is worth noting that (1) EPC++ trained with only monocular samples already outperforms \cite{godard2016unsupervised} which takes stereo pairs as input; (2) on the metrics ``Sq Rel'' and ``RMSE'', there is a large performance boost after applying the depth-flow consistency, comparing the row ``EPC++ (depth only)'' and ``EPC++ (mono depth consist)''. The two metrics measures the square of depth prediction error, and thus are sensitive to points where the depth predictions are further away from the ground truth. Applying the depth-flow consistency eliminates some ``outlier'' depth predictions. \chenxu{When we further add stereo images for training, EPC++ achieves larger performance boost compared with its monocular counter-part}. Our observation on this is that without the scale ambiguity issue in monocular training, EPC++ trained with stereo pairs benefits more from the modeling of motion segmentation and occlusions. \yang{Qualitative results of EPC++ (stereo) and EPC++ (mono) are presented in Fig. \ref{fig:stereo_examples} and \figref{fig:mono_examples} respectively. Compared to other SoA results from \cite{yang2018cvpr,godard2016unsupervised}, Our depth results preserve the details of the scene noticeably better (white circles). The green circle show in Fig. \ref{fig:mono_examples} visualizes the motion confusion discussed in Sec. \ref{sec:motion_confusion}.}

\noindent\textbf{Generalization to non-driving scenes}
To evaluate the generalization ability of our model, first, we directly apply our model trained only on the KITTI dataset to the Make3D dataset (~\cite{saxena2006learning,saxena2009make3d}), which is unseen during the training time. The comparison with other unsupervised methods is presented in Tab. \ref{table:make3d}. EPC++ trained with KITTI generalizes well to dataset with unseen scenes, outperforming other methods (\cite{zhou2017unsupervised,godard2016unsupervised}) trained with Cityscapes and KITTI datasets combined, demonstrating the effectiveness of our generalization capability. Qualitative results on Make3D dataset are shown in \figref{fig:make3d_depth}.

We also finetune our model on the MPI-Sintel Final dataset~\cite{Butler:ECCV:2012}. As only the ground truth of training split is publicly available, we report the depth evaluation results on the training set, following the traditions of optical flow evaluation on MPI-Sintel \cite{wang2017occlusion,ren2017unsupervised,Meister:2018:UUL}. Two variants of EPC++ are compared: EPC++ (mono depth only) and EPC++ (mono). An improvement from 0.866 to 0.524 in $Abs Rel$, and from 25.558 to 5.3206 in $Sq Rel$ is observed when we add depth-flow consistency, $\ve{S}$, $\ve{V}$ and adaptive training of the framework. Qualitative results are shown in upper part of \figref{fig:MPI-Sintel_depth_flow}. Please note that these results are generated by pre-trained the full EPC++ (mono) model on KITTI 2015 and finetuned on MPI-Sintel ``final'' pass, which only contains 1,000 frame pairs.
\begin{table}[!htbp]
\centering
\fontsize{8}{10}\selectfont
\def\arraystretch{1.15}
\caption{Generalization to Make3D dataset.}
\setlength{\tabcolsep}{3pt}
\begin{tabular}{lccccc}

\specialrule{.2em}{.1em}{.1em}
\multirow{2}{*}{Method} & Training & \multicolumn{4}{c}{Error Metrics}   \\ \cline{3-6}
                        &              Data              & Abs Rel & Sq Rel & RMSE  & RMSE log \\ \hline
\multicolumn{1}{l}{Godard~\etal \cite{godard2016unsupervised}}                & \chenxu{CS}                      & 0.544   & 10.94  & 11.74 & 0.193    \\
\multicolumn{1}{l}{SfMLearner~\cite{zhou2017unsupervised}}              & CS+K                      & 0.383   & 5.32   & 10.47 & 0.478    \\ \hline
\multicolumn{1}{l}{DDVO~\cite{wang2018learning}} & K & 0.387 & 4.72  & 8.09 & 0.204 \\
\multicolumn{1}{l}{Godard~\etal~\cite{godard2018digging}} & K & 0.361 & 4.17& 7.82 & 0.175 \\ \hline
\multicolumn{1}{l}{EPC++ (mono depth only)}          & K                         & 0.374   & 4.60   & 8.17  & 0.414    \\
\multicolumn{1}{l}{EPC++ (mono)}                & K                         & 0.368   & 4.22   & 7.87  & 0.409    \\
\multicolumn{1}{l}{EPC++ (stereo depth only)}   & K &  0.346 & 3.97 & 7.70 & 0.395 \\
\multicolumn{1}{l}{EPC++ (stereo)}  & K & \textbf{0.341} & \textbf{3.86} & \textbf{7.65} & \textbf{0.392} \\ \hline
\end{tabular}
\label{table:make3d}
\end{table}

\subsection{Optical Flow Evaluation}

\label{of_exp}
\noindent\textbf{Experiment setup.}
The optical flow evaluation is performed on KITTI 2015 and KITTI 2012 datasets. For ablation study, the comparison of our full model and other variants is evaluated on the training split, which consists of 200 image pairs and the ground truth optical flow is provided. We chose the training split for ablation study as the ground truth of the test split is withheld and there is a limit of submission times per month. For our full model and comparison with the SoTA methods, we evaluated on the test split and report numbers generated by the test server.

\noindent\textbf{Ablation study.}
 The ablation study our model and 4 different variants is presented in Tab. \ref{tbl:flow-ablation}.  
 The model variants include:

\noindent(1) Flow only: OptFlowNet trained with only view synthesis and smoothness losses $\hua{L}_{fvs}+\hua{L}_{fs}$.

\noindent(2) Joint training with depth: OptFlowNet is finetuned jointly with DepthNet after individually trained using $\hua{L}_{dmc}$. We can see that the results are worse than training with flow alone; this is because the flows from depth at rigid regions, \ie~$p_{st}$ in \equref{eqn:dmc}, are not as accurate as those from learning OptFlowNet alone. In other words, factorized depth and camera motion in the system can introduce extra noise to 2D optical flow estimation (from 3.66 to 4.00). But we notice that the results on occluded/non-visible regions are slightly better (from 23.07 to 22.96).

\noindent(3) EPC++ all region: We fix DepthNet, but finetune OptFlowNet without using the visibility mask $\ve{V}$. We can see the flows at rigid regions are even worse for the same reason as above, while the results at the occluded region becomes much better (from 23.07 to 16.20).

\noindent(4) EPC++ vis-rigid region: We fix DepthNet, and finetune OptFlowNet at the pixels of the visible and rigid regions, where the effect of improving at occluded region is marginal.

\noindent(5) EPC++ non-vis region: We only finetune OptFlowNet with $\hua{L}_{fc}$ and it yields improved results at all the regions of optical flow.

Results from variants (1)-(5) validate our assumption that the rigid flow from depth and camera motion helps the optical flow learning at the non-visible/occluded region. 
Comparing EPC++ vis-rigid and EPC++ non-vis for both stereo and monocular training setups, there is a large performance boost for both optical flow in occlusion regions ($occ$) and overall regions ($all$). This proves that explicitly modeling occlusion and motion mask benefits the optical flow estimation a lot.
\begin{table}[!h]
\centering
\caption{Ablation study of optical flow estimation on KITTI 2015 training set. }
\label{tbl:flow-ablation}
\def\arraystretch{1.15}
\begin{tabular}{l|ccc}
\specialrule{.2em}{.1em}{.1em}
Method & noc & occ & all \\ \hline
Flow only  & \textbf{3.66}  & 23.07 &  7.07 \\
Joint training w/ depth & 4.00 & 22.96 & 7.40 \\
EPC++ all region & 4.33  &  16.20 & 6.46 \\
EPC++ vis-rigid region &  3.97 & 21.79 & 7.17 \\
EPC++ non-vis region  &3.84  &  15.72 &  5.84 \\ \hline
EPC++ (stereo) vis-rigid region & 3.97 & 21.86 & 7.14 \\
EPC++ (stereo) non-vis region & 3.83  & \textbf{13.53}  & \textbf{5.43}  \\ \hline
\end{tabular}
\end{table}

\begin{table*}[]
\centering
\caption{Comparison of optical flow performances between EPC++ and current unsupervised SoTA on KITTI 2012 and KITTI 2015 datasets. The numbers reported on KITTI 2012 dataset use average EPE metric. The numbers reported on KITTI 2015 train split use EPE metric while the numbers reported on test split use the percentage of erroneous pixels (F1 score) as generated by the KITTI evaluation server. \yang{Network architectures are specified in parentheses.}}
\label{tbl:of-sota}
\def\arraystretch{1.15}
\begin{tabular}{l|l|r|clc|clccc}
\specialrule{.2em}{.1em}{.1em}

\multirow{3}{*}{Method} & \multirow{3}{*}{Backbone} &
\multirow{3}{*}{\#params} & \multicolumn{3}{c|}{KITTI 2012}                              & \multicolumn{5}{c}{KITTI 2015}                                                                \\
          &&              & Train         &                               & Test         & Train         &                      & \multicolumn{3}{c}{Test}                               \\ \cline{4-4} \cline{6-7} \cline{9-11}
                  & &      & all           &                               & all          & all           &                      & bg               & fg               & all              \\ \hline
DSTFlow \cite{ren2017unsupervised} & FlowNet-C &   39.2M           & 10.43         &                               & 12.40        & 16.79         &                      & -                & -                & 39.00\%          \\
OccAwareFlow~\cite{wang2017occlusion}  &FlowNet-C&     39.2M         & 3.55          &                               & 4.20         & 8.88          &                      & -                & -                & 31.20\%          \\
Unflow-CSS \cite{Meister:2018:UUL} & FlowNet-CSS&     116.6M        & 3.29          &                               & -            & 8.10          &                      & -                & -                & -                \\
Multi-frame~\cite{janai2018unsupervised}  &multiview PWC-Net &          & -             &                               & -            & 6.59             &                      & 22.67\%          & \textbf{24.27\%}          & 22.94\%          \\ \hline
GeoNet \cite{yin2018geonet}    & ResNet-50 &   58.5M           & -             &                               & -            & 10.81         &                      & -                & -                & -                \\
DF-Net \cite{zou2018dfnet}   & FlowNet-C  &   39.2M                 & 3.54          &                               & 4.40         & 8.98          &                      & -                & -                & 25.70\%          \\
Competitive-Collaboration-uft~\cite{ranjan2018adversarial}   & PWC-Net  & 5.1M  & -             &                               & -            & 5.66             &                      & -          & -          & 25.27\%          \\ \hline
EPC++ (mono)   & PWC-Net  &   5.1M           & 2.30 & \multicolumn{1}{c}{\textbf{}} & 2.60 & 5.84          & \multicolumn{1}{c}{} & 20.61\% & 26.32\% & 21.56\% \\
EPC++ (stereo)   & PWC-Net  &  5.1M        & \textbf{1.91}             & \multicolumn{1}{c}{}          & \textbf{2.20}            & \textbf{5.43} & \multicolumn{1}{c}{} & \textbf{19.24\%}               & 26.93\%                & \textbf{20.52\% }               \\ \hline
\end{tabular}
\end{table*}

\begin{figure*}
\centering
\includegraphics[width=\textwidth]{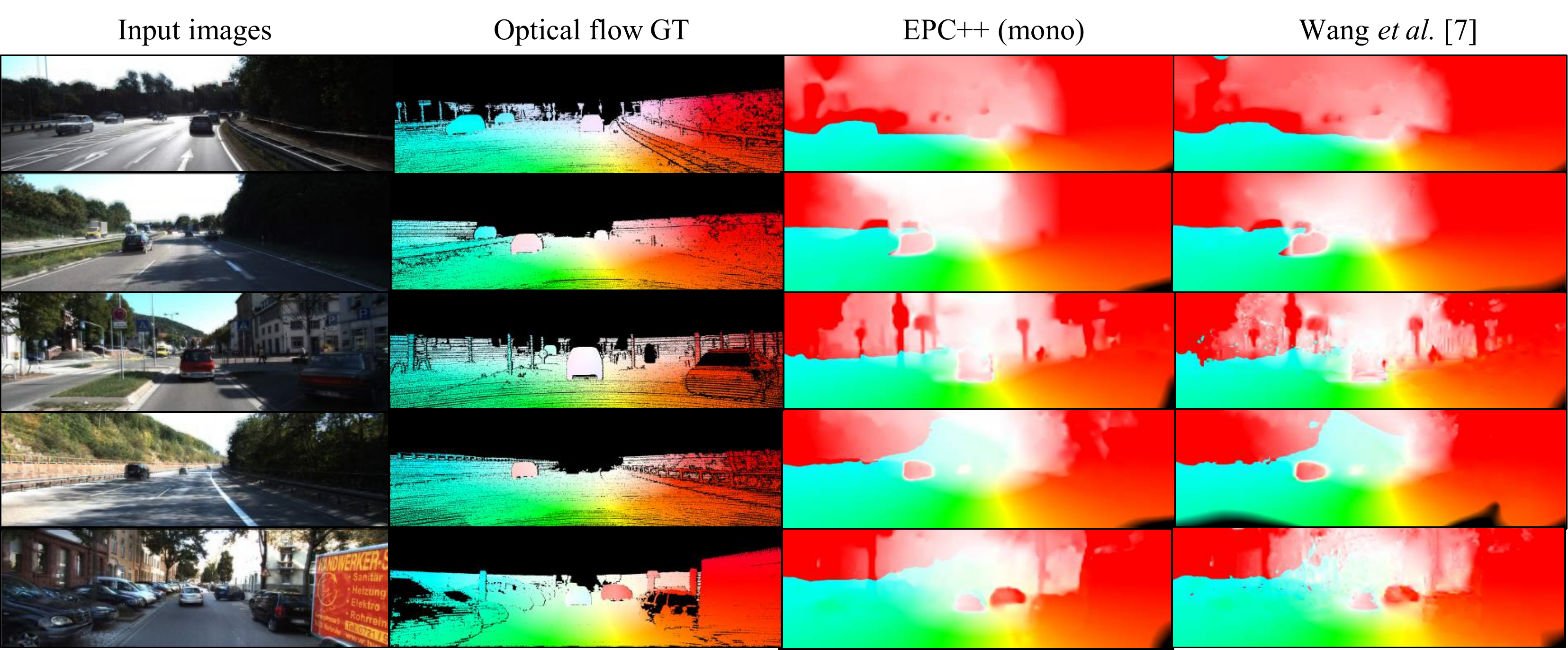}
\caption{Visualization of optical flow results on KITTI 2015 training set images. We compared with current unsupervised SoTA method \cite{wang2017occlusion}. Optical flow results generated by EPC++ align better with the ground truth, especially on object boundaries (occlusion regions).}
\label{fig:flow}
\end{figure*}

\noindent\textbf{Comparison with SoTA methods.}
For fair comparison with current SoTA optical flow methods, our OptFlowNet is evaluated on both KITTI 2015, KITTI 2012 training and test splits. On test split, the reported numbers are generated by official evaluation servers. As shown in the Tab. \ref{tbl:of-sota}, EPC++ (mono) outperforms all current unsupervised monocular methods (\cite{ren2017unsupervised,Meister:2018:UUL,wang2017occlusion,yin2018geonet,zou2018dfnet}) on ``F1-bg'' and ``F1-all'' metrics. Please note that Multi-frame \cite{janai2018unsupervised} reports better performance on ``F1-fg'', but this method takes three frames as input to estimate the optical flow while our method only takes two. Although longer input sequence gives better estimation the movement of foreground objects, our results at full regions are still better. \yang{EPC++ (stereo) shows a further performance boost compared to monocular counterpart. This is as expected as better depth estimation provides better guidance for optical flow training.}

Comparing our method with DF-Net \cite{zou2018dfnet}, whose optical flow network is firstly pre-trained on additional SYNTHIA dataset \cite{ros2016synthia} and then jointly trained on KITTI dataset, we have achieved a large performance gain on both KITTI 2012 and KITTI 2015 datasets. Contrast to DF-Net, our method also models the pixels in occlusion regions and adopt an adaptive training to leverage depth information to help optical flow learning. Qualitative results are shown in Fig.\ref{fig:flow}, and ours have better sharpness and smoothness of the optical flow.
\vspace{-0.1\baselineskip}

\noindent\textbf{Generalization to MPI-Sintel Dataset.}
MPI-Sintel~\cite{Butler:ECCV:2012} is a synthetic benchmark used for optical flow evaluation. It provides very different scenes compared with KITTI. To better compare with previous works \cite{ren2017unsupervised,wang2017occlusion,Meister:2018:UUL}, we have adopted two training setups: (1) test our model (which is trained on KITTI 2015) directly on MPI-Sintel data (Trained on KITTI); (2) the pre-trained model (on KITTI) is further finetuned with MPI-Sintel training data (Fine-tuned on MPI-Sintel). We apply the same parameters and training strategy as used on KITTI.

\begin{figure*}
\centering
\includegraphics[width=\textwidth]{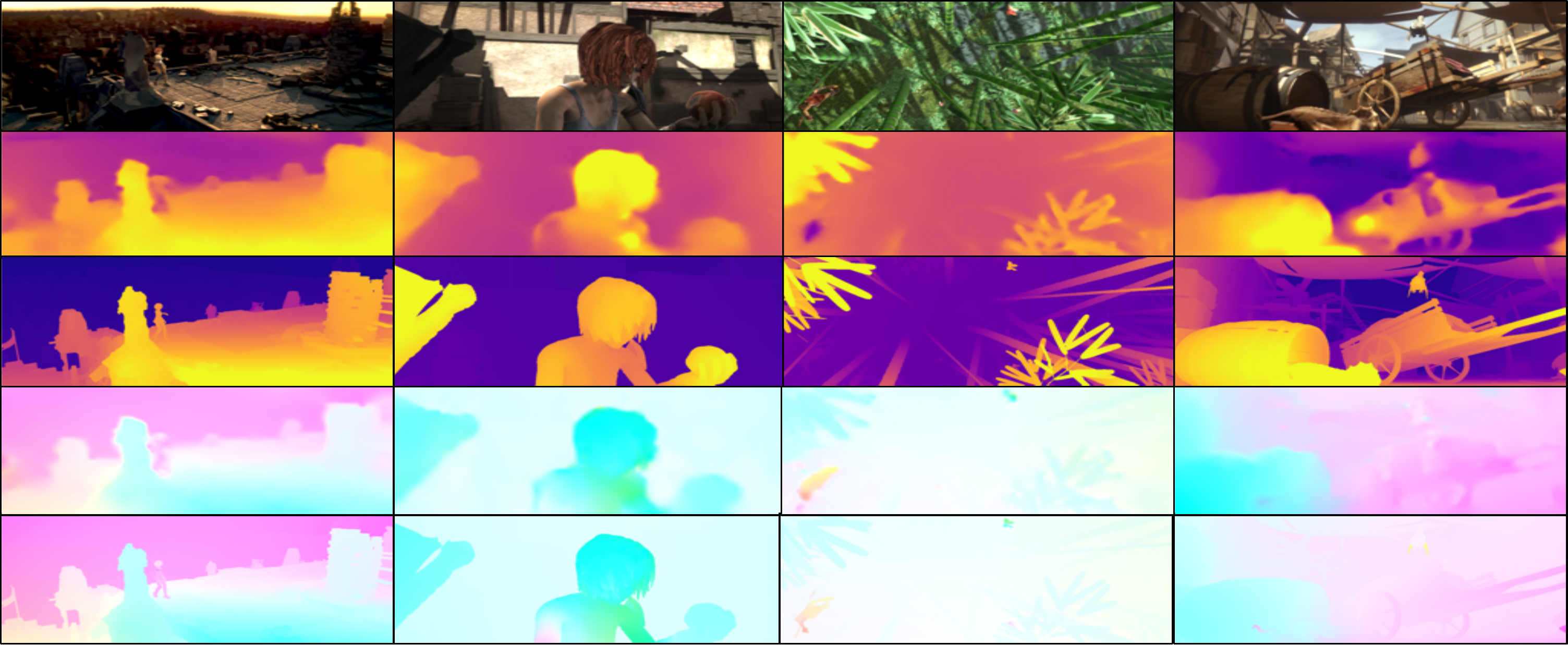}
\caption{Visualization of depth and optical flow estimation results on MPI-Sintel dataset using our model fine-tuned on MPI-Sintel. From top to bottom: input image, depth estimation result, depth ground truth, optical flow estimation result, optical flow ground truth.}
\label{fig:MPI-Sintel_depth_flow}
\end{figure*}

From the results shown in Table \ref{tbl:MPI-Sintel}, we can see that the results are consistent with the ablation study on the KITTI dataset, whether tested directly or fine-tune on the MPI-Sintel dataset.  This shows that our proposed training schedule can also generalize well to other scenarios.
\begin{table}[]
\centering
\caption{Optical flow performance of unsupervised methods on the MPI-Sintel final training split. We report the EPE metrics using model trained only on the KITTI dataset in the left part and present models fine-tuned on the MPI-Sintel dataset in the right part. }
\label{tbl:MPI-Sintel}
\setlength{\tabcolsep}{8pt}
\def\arraystretch{1.15}
\begin{tabular}{lcc}
\specialrule{.2em}{.1em}{.1em}
Method                              & Trained on KITTI & Fine-tuned on the MPI-Sintel \\ \hline
\multicolumn{1}{l|}{DSTFlow~\cite{ren2017unsupervised}} & 7.95             & 6.81                            \\
\multicolumn{1}{l|}{OccAware~\cite{wang2017occlusion}}  & 7.92             & 5.95                            \\
\multicolumn{1}{l|}{Unflow-CSS~\cite{Meister:2018:UUL}} & -                & 7.91                            \\ \hline
\multicolumn{1}{l|}{EPC++ flow only}                    & 7.33             & 5.90                            \\
\multicolumn{1}{l|}{EPC++ all region}                  & 6.95             & 6.21                            \\
\multicolumn{1}{l|}{EPC++}                              & \textbf{6.67}    & \textbf{5.64} \\ \hline
\end{tabular}
\end{table}
The qualitative results are shown in bottom part of \figref{fig:MPI-Sintel_depth_flow}.

\begin{figure*}
\centering
\includegraphics[width=1.0\textwidth]{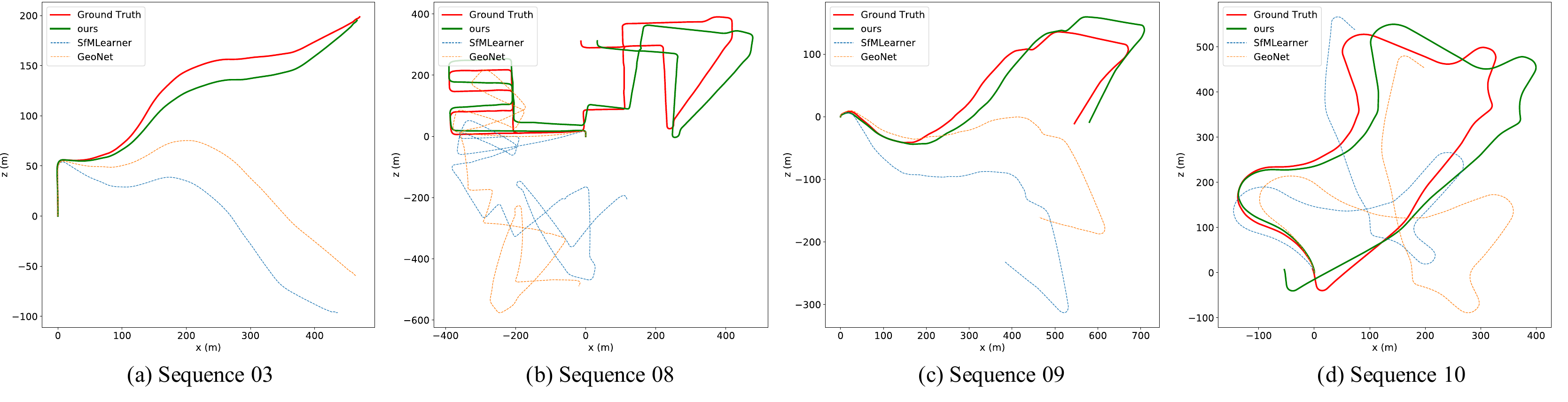}
\caption{Odometry estimation results on four sequences of KITTI 2015 dataset. The two left figures (a) and (b) are results on training sequences and the right two results on test sequences.}
\label{fig:odo}
\end{figure*}

\subsection{Odometry estimation.}
To evaluate the performance of our trained MotionNet, we use the odometry metrics as in \cite{zhou2017unsupervised,yin2018geonet}. The same protocol as in \cite{zhou2017unsupervised} is applied in our evaluation, which measures the absolute trajectory error averaged every five consecutive frames.
Unlike the settings in previous works~\cite{zhou2017unsupervised,yin2018geonet} which train a MotionNet using stacked five frames (as described in Sec. \ref{sec:approach}), no modifications have been made to the MotionNet, which still takes three frames as input and retrain our networks on KITTI odometry train split.
We compare with several unsupervised SoTA methods on two sequences of KITTI odometry test split. To explore variants of our model, we experimented learning with monocular samples (EPC++ (mono)) and with stereo pairs (EPC++ (stereo)).

As shown in Table \ref{table:odo}, our trained MotionNet shows superior performance with respect to visual SLAM methods (ORB-SLAM), and is comparable to other unsupervised learning methods with slight improvement on two test sequences. The more accurate depth and optical flow estimation from our DepthNet helps constraint the output of MotionNet, yielding better odometry results.
The qualitative odometry results are shown in Fig. \ref{fig:odo}. Compared to results from SfMLearner \cite{zhou2017unsupervised} or GeoNet \cite{yin2018geonet}, which have large offset at the end of the sequence, results  from EPC++are more robust to large motion changes and closer to the ground truth trajectories.

\begin{table}[]
\centering
\caption{Odometry evaluation on two sequences of KITTI 2015 dataset. All presented results are generated by unsupervised methods.}
\label{table:odo}
\setlength{\tabcolsep}{5pt}
\def\arraystretch{1.2}
\begin{tabular}{lcc}
\specialrule{.2em}{.1em}{.1em}
Method & Seq.09 & Seq.10 \\ \hline
\multicolumn{1}{l|}{ORB-SLAM (full) \cite{mur2015orb}} &  $0.014\pm 0.008$ & $0.012\pm 0.011$ \\
\multicolumn{1}{l|}{ORB-SLAM (short) \cite{mur2015orb}} &  $0.064\pm 0.141$ & $0.064\pm 0.130$ \\
\multicolumn{1}{l|}{Zhou \etal ~\cite{zhou2017unsupervised}} &  $0.021\pm0.017$  & $0.020\pm0.015$ \\
\multicolumn{1}{l|}{DF-Net~\cite{zou2018dfnet}} &  $0.017\pm0.007$ &  $0.015\pm0.009$ \\
\multicolumn{1}{l|}{Mahjourian\etal~\cite{mahjourian2018unsupervised}} &  $0.013\pm0.010$ &  $0.012\pm0.011$ \\
\multicolumn{1}{l|}{GeoNet~\cite{yin2018geonet}} &  \textbf{0.012 $\pm$0.007} &  \textbf{0.012$\pm$0.009} \\ \hline
\multicolumn{1}{l|}{EPC++(mono)} & $0.013\pm0.007$  & \textbf{0.012 $\pm$ 0.008}\\
\multicolumn{1}{l|}{EPC++(stereo)} & \textbf{0.012 $\pm$ 0.006} & \textbf{0.012 $\pm$ 0.008} \\\hline
\end{tabular}
\end{table}

\begin{table}[]
\centering
\caption{Odometry evaluation on KITTI dataset using the metric of average translation and rotation errors.}
\label{tbl:odo_full}
\setlength{\tabcolsep}{5pt}
\def\arraystretch{1.15}
\begin{tabular}{l|cc|cc}
\specialrule{.2em}{.1em}{.1em}
\multirow{2}{*}{Method}               & \multicolumn{2}{c|}{Seq. 09}                                              & \multicolumn{2}{c}{Seq. 10}                                               \\
                                       & $t_{err}\%$ & $r_{err}(^{\circ}/100)$ & $t_{err}\%$ & $r_{err}(^{\circ}/100)$ \\ \hline
Zhou \etal \cite{zhou2017unsupervised} & 30.75        & 11.41                                                       & 44.22        & 12.42                                                      \\
GeoNet \cite{yin2018geonet}        & 39.43        & 14.30                                                      & 28.99        & 8.85                                                            \\ \hline
EPC++ (mono)                            & \textbf{8.84}   & \textbf{3.34}                                           & \textbf{8.86}        & \textbf{3.18}                                                       \\ \hline
\end{tabular}
\end{table}

\begin{figure*}
\centering
\includegraphics[width=\textwidth]{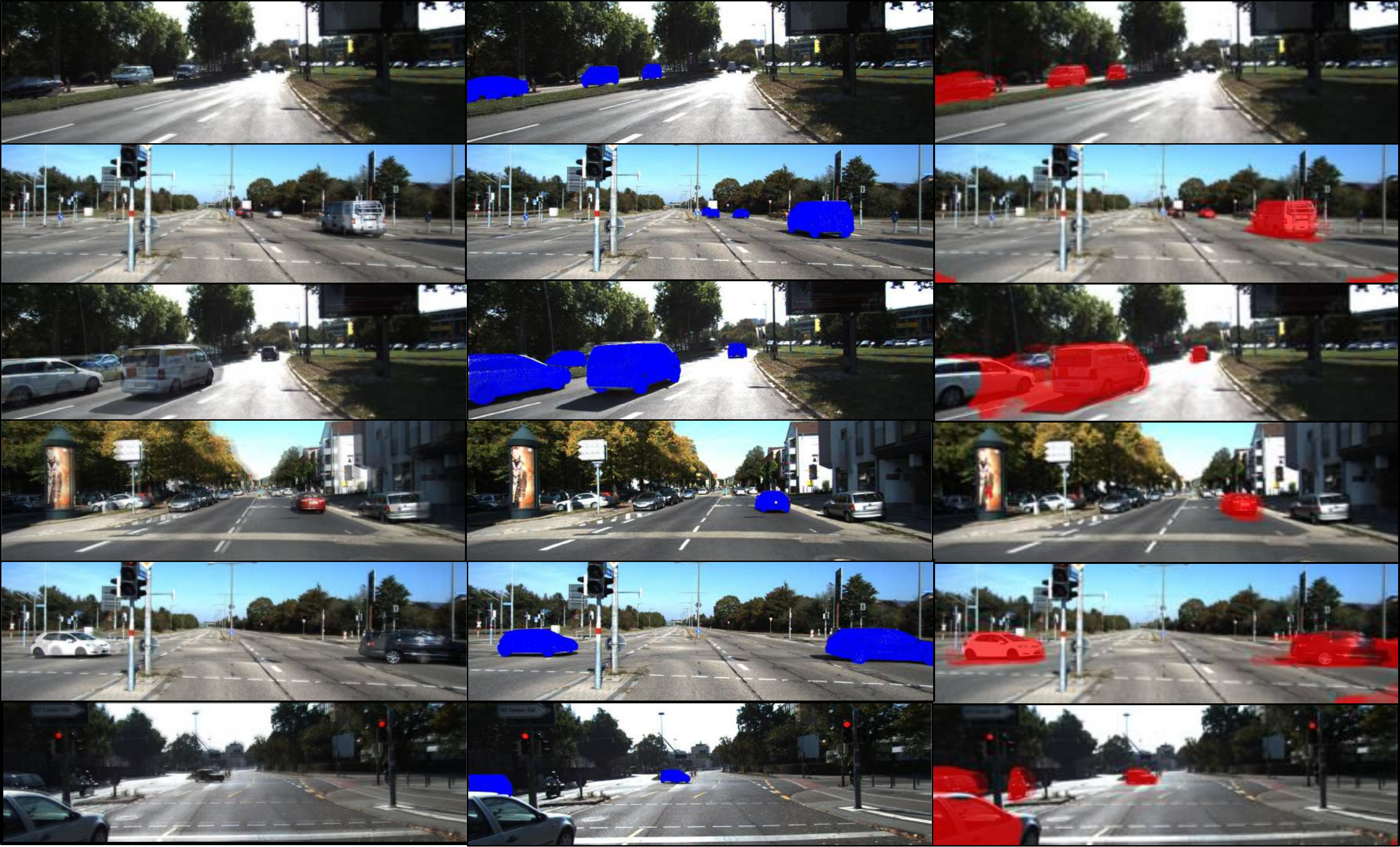}
\caption{Moving object segmentation results on KITTI training split. \chenxu{The ground truth masks are shown in blue and the red ones are our predictions. }}
\label{fig:seg}
\end{figure*}

The small quantitative performance gap leads to large qualitative performance difference because the metric only evaluates 5-frame relative errors and always assume the first frame prediction to be ground truth; thus the errors can add up in the long test sequence while the existing metrics do not take it into consideration.
To better compare the odometry performance over the complete sequence, we adopted the evaluation metrics as proposed in \cite{li2017undeepvo}. This metric evaluates the average translational and rotational errors over the full sequence and the quantitative results are shown in Tab. \ref{tbl:odo_full}. As these metrics evaluate over the full sequence, the quantitative numbers align well with the qualitative results in Fig. \ref{fig:odo}. In summary, by applying the same MotionNet architecture as in EPC++ pipeline on KITTI odometry split, we have achieved SoTA performance on standard evaluation metric. On a metric which focuses on the long-term odometry accuracy, EPC++ outperforms the previous works by a large margin. As the ego-motion is coupled and jointly trained with depth and optical flow, the performance boost of depth and optical flow help regularize the learning of odometry.

\subsection{Moving object segmentation}
\label{seg_exp}
Ideally, the residual between the dynamic scene flow $\ve{M}_d$ and the background scene flow $\ve{M}_b$ represents the motion of foreground object. As the HMP (\equref{eqn:hmp}) is capable of decomposing the foreground and background motion by leveraging the depth-flow consistency, we test the effectiveness of this decomposition by evaluating the foreground object segmentation.

\noindent\textbf{Experiment setup.} The moving object segmentation is evaluated on the training split of the KITTI 2015 dataset. An ``Object map'' is provided in this dataset to distinguish the foreground and background in flow evaluation. We use this motion mask as ground truth in our segmentation evaluation. Fig. \ref{fig:seg} (second column) shows some visualizations of the segmentation ground truths. Our foreground segmentation estimation is generated by subtracting the rigid optical flow from optical flow, as indicated by $\ve{S}$ in \equref{eqn:hmp}. We set a threshold on $\ve{M}_d > 3$ to generate a binary segmentation mask. 

\begin{table}[t]
\centering
\setlength{\tabcolsep}{3pt}
\caption{Foreground moving object segmentation performance on KITTI 2015 dataset.}
\label{tbl:seg}
\setlength{\tabcolsep}{2pt}
\def\arraystretch{1.15}
\begin{tabular}{l|cccc}
\specialrule{.15em}{.08em}{.08em}
 Method                            & pixel acc. & mean acc. & mean IoU & f.w. IoU \\ \hline
Explainability mask \cite{zhou2017unsupervised} & 0.61      & 0.54     & 0.38    & 0.64    \\
EPC (stereo) \cite{yang2018every}         & 0.89                & 0.75                & 0.52              & \textbf{0.87}             \\
Graphcut on residual (stereo) & 0.76 & 0.46 & 0.40 & 0.78 \\\hline
EPC++(mono) & 0.88 & 0.63 & 0.50 & 0.86 \\
EPC++(stereo)                         & \textbf{0.91}      & \textbf{0.76}     & \textbf{0.53}    & \textbf{0.87}   \\ \hline
\end{tabular}
\end{table}

\begin{table*}[!htpb]
\centering
\caption{Scene flow performances of different methods on KITTI 2015 training split.}
\label{tbl:sf_sota}
\setlength{\tabcolsep}{2pt}
\def\arraystretch{1.25}
\begin{tabular}{l|c|c|ccc|ccc|ccc}
\specialrule{.2em}{.1em}{.1em}
\multirow{2}{*}{Method} & \multirow{2}{*}{Test data} & \multirow{2}{*}{Supervision} & \multicolumn{3}{c|}{D1}                          & \multicolumn{3}{c|}{D2}                          & \multicolumn{3}{c}{FL}                           \\
          & &              & bg             & fg             & bg+fg          & bg             & fg             & bg+fg          & bg             & fg             & bg+fg          \\ \hline
OSF \cite{menze2015cvpr}    & partial   & yes              & 4.00           & 8.86           & 4.74           & 5.16           & 17.11          & 6.99           & 6.38           & 20.56          & 8.55 \\
ISF \cite{behl2017bounding}   & partial   & yes           & 3.55            & 3.94          & 3.61           & 4.86           & 4.72           & 4.84           & 6.36           & 7.31           & 6.50       \\ \hline
\multicolumn{10}{c}{} \\ \hline
EPC (stereo) \cite{yang2018every}  & full  & no      & 23.62          & 27.38          & 26.81          & 18.75          & 70.89          & 60.97          & 25.34          & 28.00          & 25.74          \\ \hline
EPC++ (mono)   & full  & no          & 30.67          & 34.38          & 32.73          & 18.36          & 84.64          & 65.63          & \textbf{17.57} & 27.30          & 19.78          \\
EPC++ (stereo)   & full  & no        & \textbf{22.76} & \textbf{26.63} & \textbf{23.84} & \textbf{16.37} & \textbf{70.39} & \textbf{60.32} & 17.58          & \textbf{26.89} & \textbf{19.64} \\ \hline
\end{tabular}
\end{table*}
\noindent\textbf{Evaluation results.}
The quantitative and qualitative results are presented in Tab. \ref{tbl:seg} and Fig. \ref{fig:seg} respectively. We compare with two previous methods \cite{zhou2017unsupervised,yang2018every} that takes the non-rigid scene into consideration.
Yang \etal~\cite{yang2018every}
explicitly models the moving object mask, and thus is directly comparable. The ``explainability mask'' in \cite{zhou2017unsupervised} is designed to deal with both moving objects and occlusion, and here we list their performances for a more comprehensive comparison.  Our generated foreground segmentation outperforms the previous methods on all metrics, and the visualization shows the motion mask aligns well with the moving object. It is worth noting that monocular EPC++, despite scale ambituity issue, already performs comparable to EPC \cite{yang2018every}, which is trained with stereo pairs. This proves the effectiveness of the modeling of segmentation mask in our pipeline.

\subsection{Scene flow evaluation}
\label{sf_exp}
\noindent\textbf{Experiment setup.}
The scene flow evaluation is performed on training split of KITTI 2015 dataset. There are 200 frames pairs (frames for $t$ and $t+1$) in the scene flow training split. The depth ground truth of the two consecutive frames and the 2D optical flow ground truth from frame $t$ to frame $t+1$ are provided. The evaluation of scene flow is performed with the KITTI benchmark evaluation toolkit~\footnote{\url{http://www.cvlibs.net/datasets/kitti/eval_scene_flow.php}}.
As the unsupervised monocular method generates depth/disparity without absolute scale, we rescale the estimated depth by matching the median to ground truth depth for each image.
Since no unsupervised methods have reported scene flow performances on KITTI 2015 dataset, we compare our model trained on monocular sequences (EPC++ (mono)) and stereo pairs (EPC++ (stereo)) with the previous results reported in \cite{yang2018every}. As shown in Tab. \ref{tbl:sf_sota}, our scene flow performance outperforms the previous SoTA method \cite{yang2018every}. Although it is not completely fair comparison, the performances of OSF \cite{menze2015cvpr} and ISF \cite{behl2017bounding} are also presented in the results table. Both methods are supervised and the performances are generated by training the model on part of the training set and evaluated on the rest.


\section{Conclusion} 
In this paper, we presented an end-to-end unsupervised learning framework, which we call every pixel counts ++ (EPC++), for jointly estimating depth, camera motion, optical flow and moving object segmentation masks. It successfully leverages the benefits of different tasks by exploiting their geometric consistency.

In our framework, we proposed and adopted a depth, ego-motion and optical flow consistency with explicit awareness of both motion rigidity and visibility. Thus every pixel can be explained by either rigid motion, non-rigid/object motion or occluded/non-visible regions. We proposed an adaptive training strategy to better leverage the different advantages of depth or optical flow and showed better performance than directly applying uniform across-task consistency.

We conducted comprehensive experiments to evaluate the performance of EPC++ over different datasets, and showed SoTA performance on both driving scenes (KITTI) and non-driving scenes (Make3D, MPI-Sintel) over all the related tasks. This demonstrates the effectiveness and also good generalization capability of the proposed framework. 
In the future, we hope to apply EPC++ to other motion videos containing deformable and articulated non-rigid objects such as the ones from MoSeg~\cite{brox2010object} \etc, and extend EPC++ to multiple object segmentation, which provides object part and motion understanding in an unsupervised manner.
\ifCLASSOPTIONcaptionsoff
  \newpage
\fi




\bibliographystyle{IEEEtran}
\bibliography{IEEEabrv,unsp_motion_depth}




%

\begin{IEEEbiography}[{\includegraphics[width=1in,height=1.25in,clip,keepaspectratio]{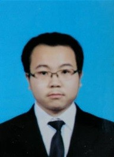}}]{Chenxu Luo}
Chenxu Luo is currently a Ph.D student in Computer Science at Johns Hopkins University. He received the BS degree in Information and Computing Science from Schools of Mathematical Sciences, Peking University in 2016. His research interests include computer vision and robotics.
\end{IEEEbiography}

\begin{IEEEbiography}[{\includegraphics[width=1in,height=1.25in,clip,keepaspectratio]{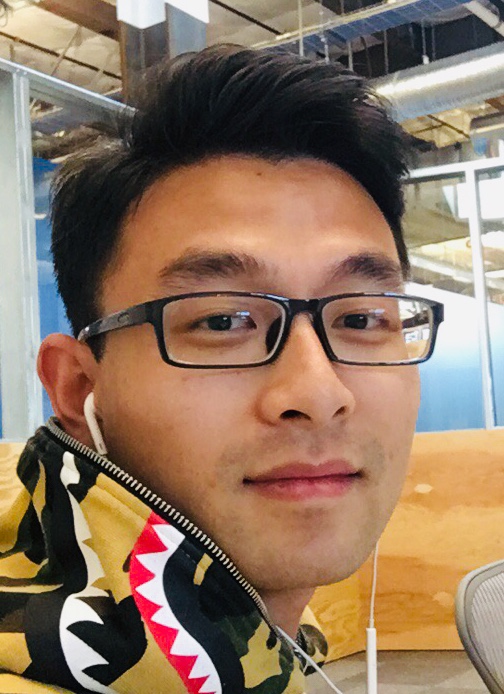}}]{Zhenheng Yang}
is currently a research scientist at Facebook AI. He obtained his PhD degree in University of Southern California under supervision of Prof. Ram Nevatia. He received his BEng degree from Tsinghua University in 2014. His research interests include computer vision and machine learning. He has published in top computer vision conferences like CVPR, ICCV, ECCV, AAAI, etc.
\end{IEEEbiography}

\begin{IEEEbiography}[{\includegraphics[width=1in,height=1.25in,clip,keepaspectratio]{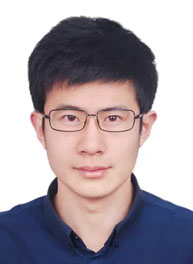}}]{Peng Wang}
is a research scientist in Baidu USA LLC. He obtained his Ph.D. degree in University of California, Los Angeles, advised by Prof. Alan Yuille. Before that, he received his B.S. and M.S. from Peking University, China. His research interest is image parsing and 3D understanding, and vision based autonomous driving system. He has around 20 published papers in ECCV/CVPR/ICCV/NIPS.
\end{IEEEbiography}

\begin{IEEEbiography}[{\includegraphics[width=1in,height=1.25in,clip,keepaspectratio]{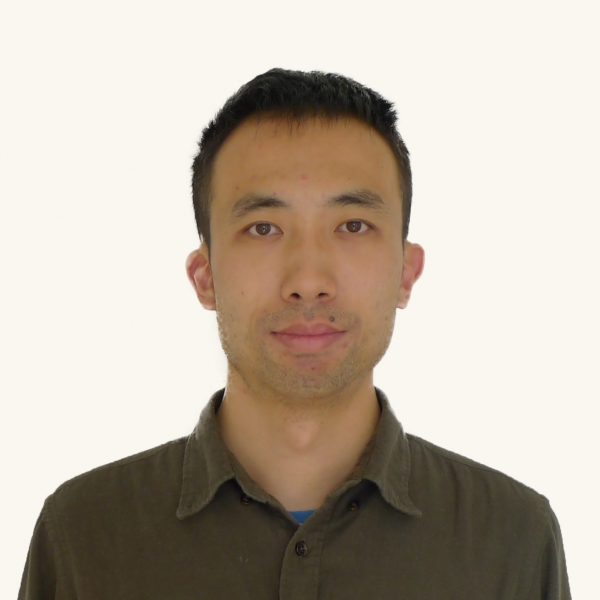}}]{Yang Wang}
Yang Wang is a research scientist in Baidu USA. He obtained his Ph.D. degree in physics at UC Berkeley. Before that, he received B.S. degree from Tsinghua University in China. His research interest lies in unsupervised learning methods in computer vision problems.
\end{IEEEbiography}

\begin{IEEEbiography}[{\includegraphics[width=1in,height=1.25in,clip,keepaspectratio]{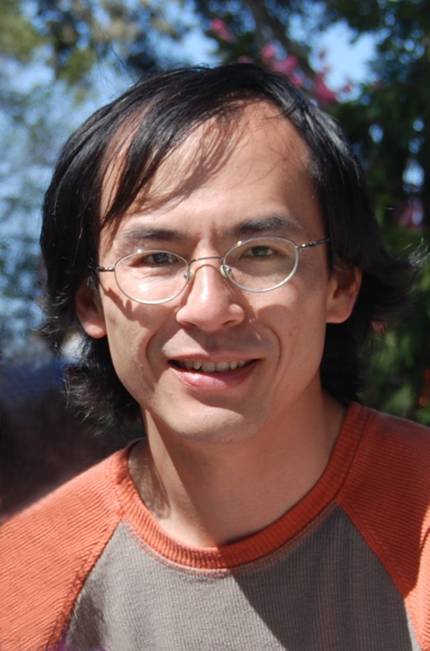}}]{Wei Xu}
Wei Xu is the chief scientist of general AI at Horizon Robotics Inc., where focuses on artificial general intelligence research. Before that, he was a distinguished scientist at Baidu Inc., where he started Baidu's open source deep learning framework PaddlePaddle and led general AI research. Before that, he was a research scientist at Facebook, where he developed and deployed recommendation system capable of handling billions of objects and billion+ users. Before that he was a senior researcher at NEC Laboratories America, where he developed convolutional neural networks for a variety of visual understanding tasks. He received his B.S. degree from Tsinghua University and his M.S. degree from Carnegie Mellon University.
\end{IEEEbiography}

\begin{IEEEbiography}[{\includegraphics[width=1in,height=1.25in,clip,keepaspectratio]{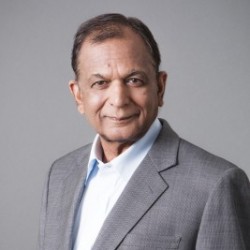}}]{Ram Nevatia}
received the PhD degree in electrical engineering from Stanford University, CA. He has been with the University of Southern California, Los Angeles, since 1975, where he is currently a professor of computer science and electrical engineering and the director of the Institute for Robotics and Intelligent Systems. He has also made contributions to the topic of object recognition, stereo analysis, modeling from aerial images, action recognition and tracking.
\end{IEEEbiography}

\begin{IEEEbiography}[{\includegraphics[width=1in,height=1.25in,clip,keepaspectratio]{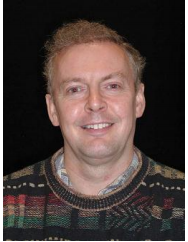}}]{Alan L. Yuille }
(F’09) received the BA degree in
math- ematics from the University of Cambridge
in 1976. His PhD on theoretical physics, supervised
by Prof. S.W. Hawking, was approved in
1981. He was a research scientist in the Artificial
Intelligence Laboratory at MIT and the Division
of Applied Sciences at Harvard University from
1982 to 1988. He served as an assistant and
associate professor at Harvard until 1996. He
was a senior research scientist at the SmithKettlewell
Eye Research Institute from 1996 to
2002. He joined the University of California, Los Angeles, as a full
professor with a joint appointment in statistics and psychology in 2002,
and computer science in 2007. He was appointed a Bloomberg Distinguished
Professor at Johns Hopkins University in January 2016. He
holds a joint appointment between the Departments of Cognitive science
and Computer Science. His research interests include computational
models of vision, mathematical models of cognition, and artificial intelligence and neural network
\end{IEEEbiography}




\end{document}